# Generative Probabilistic Planning for Optimizing Supply Chain Networks

**Hyung-il Ahn, Santiago Olivar, Hershel Mehta, Young Chol Song**
Noodle AI
San Francisco, CA 94104
{hyungil.ahn, santiago.olivar, hershel.mehta, youngchol.song}@noodle.ai

## Abstract

Supply chain networks in enterprises are typically composed of complex topological graphs involving various types of nodes and edges, accommodating numerous products with considerable demand and supply variability. However, as supply chain networks expand in size and complexity, traditional supply chain planning methods (e.g., those found in heuristic rule-based and operations research-based systems) tend to become locally optimal or lack computational scalability, resulting in substantial imbalances between supply and demand across nodes in the network. This paper introduces a novel Generative AI technique, which we call Generative Probabilistic Planning (GPP). GPP generates dynamic supply action plans that are globally optimized across all network nodes over the time horizon for changing objectives like maximizing profits or service levels, factoring in time-varying probabilistic demand, lead time, and production conditions. GPP leverages attention-based graph neural networks (GNN), offline deep reinforcement learning (Offline RL), and policy simulations to train generative policy models and create optimal plans through probabilistic simulations, effectively accounting for various uncertainties. Our experiments using historical data from a global consumer goods company with complex supply chain networks demonstrate that GPP accomplishes objective-adaptable, probabilistically resilient, and dynamic planning for supply chain networks, leading to significant improvements in performance and profitability for enterprises. Our work plays a pivotal role in shaping the trajectory of AI adoption within the supply chain domain.

## Introduction

With the pandemic wreaking havoc on supply chain networks, companies have come to understand that their traditional supply chain planning cannot cope with the challenges posed by unpredictable situations. Amidst the "new normal," organizations confront the responsibility of navigating substantially increased demand and supply variability, as well as elevated risks of disruptions across global supply chain networks (Shih, 2020). In this environment, enterprises must create efficient and effective supply chain planning to balance demand and supply and to optimize costs.

Typical supply chain networks in enterprises involve a variety of complex topological graphs over different types of nodes and edges for a plethora of SKUs (stock keeping units) with significant uncertainties, such as demand fluctuations, lead time variations, and production schedule changes, which make it hard to predict future conditions and devise optimal supply action plans (Davis, 1993; Minner, 2003). These time-varying uncertain conditions require incorporating the underlying probabilistic characteristics of variables into supply chain planning (You and Grossmann, 2008). However, conventional heuristic rule-based and operations research (OR)-based methods for supply chain planning typically rely on deterministic fixed conditions or known probability distribution conditions (Birge and Louveaux, 2011; E. A. Silver et al., 2016; Vandeput, 2020) and create inflexible plans that become vulnerable in the face of real-world complex uncertainties (Petrovic et al., 1998). Heuristic rule-based planning methods like constructive heuristic and local search lack globally defined constraints and cost objectives, resulting in locally optimal and myopic plans (Silver, 2004). Meanwhile, OR-based planning methods like mixed-integer programming (MIP) and constraint programming (CP) tend to be computationally expensive, and often intractable, for large-scale combinatorial planning problems in complex supply chains (Denton et al., 2006; Gamrath et al., 2019; Kochenderfer and Wheeler, 2019; Rossi et al., 2008; Schewe et al., 2020). As a result, these approaches are not well-suited for dynamically re-generating plans to adapt to changing states and objectives.

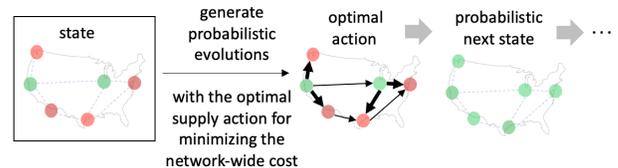

Figure 1: Generative Probabilistic Planning (GPP)

This paper introduces a novel Generative AI technique, which we call Generative Probabilistic Planning (GPP), designed to tackle these challenges in supply chain networks. GPP generates dynamic supply plans that are optimized across all network nodes over the time horizon for changing objectives like maximizing profits or service levels, factoring in time-varying probabilistic demand, lead time, and production conditions. GPP achieves these objectives by combining attention-based graph neural networks (GNN), offline deep reinforcement learning (Offline RL), and policy simulations. The power of GNN (Bronstein et al., 2021;

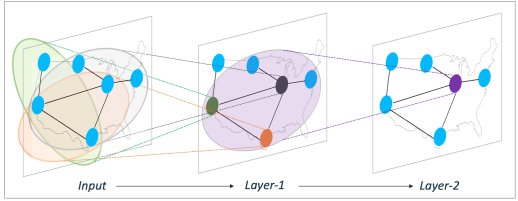

Figure 2: Graph Convolution Layers and Node Embedding.

Hamilton et al., 2017; Kipf and Welling, 2017) lies in the ability to thoroughly represent a tremendous number of dynamic and interactive patterns of the possible states and actions in supply chain network data over time, capturing complex relationships, dependencies, and constraints. We employ graph attention networks (GAT), a type of GNN with dynamic attention mechanisms that empower our model to adaptively assign varying weights to different connections, depending on the current states of the nodes and edges (Brody et al., 2022; Veličković et al., 2018). On top of that, we use Offline RL (Agarwal et al., 2020; Fujimoto and Gu, 2021; Janner et al., 2021; Kumar et al., 2020; Levine et al., 2020; Prudencio et al., 2023) to train GNN-based generative policy models with varying risk preferences[1], based on the historical network transition data that most organizations have already accumulated over years. We employ the trained risk-preferential policies in a switching or weighted ensemble manner to create optimal supply plans in probabilistic simulations that incorporate various uncertainties in demand, lead time, and production over networks for changing objectives. This enables enterprises to make objective-adaptable, probabilistically resilient, and dynamic planning for supply chain networks and strikingly improve performance and profitability.

Like the Generative AI breakthrough achieved by ChatGPT (OpenAI, 2023) in the Q&A domain, where deep attention-based sequence transformers (Vaswani et al., 2017) for large language models (LLM) were combined with reinforcement learning from human feedback (RLHF) (Christiano et al., 2017), GPP marks a pioneering Generative AI technology specifically developed for supply chain planning, combining GNN and Offline RL. GPP can *generate* probabilistic samples of the dynamic evolution of supply chain networks with supply actions that optimize costs for changing objectives, factoring in probabilistic current states and future conditions (analogous to *prompts* in ChatGPT) inferred from current inventory, predicted demand, lead time and production (Figure 1).

Our experiment results for GPP are based on the historical dataset obtained from a global consumer goods company with complex supply chain networks, containing weekly network snapshots of 500 high-volume SKUs. Using counterfactual simulations with a trained GPP policy on a hold-out testing dataset, we contrasted the supply actions generated by GPP with the historical actions taken by the enterprise. Remarkably, GPP revealed a 75% reduction in lost sales, leading to increased revenue and margin, as well as a 20% reduction in excess stocks, resulting in decreased inventory holding cost. In summary, our contributions can be highlighted in two aspects.

- We present GPP, a novel Generative AI technique tailored for planning in supply chain networks. Our methodology employs GNN to represent complex relationships and patterns in supply chain data. GNN-based policies for varying risk preferences are simultaneously pre-trained using Offline RL, compared, and selected in probabilistic simulations to generate optimal supply plans for cost objectives.
- We conduct experiments with real-world large-scale complex supply chain data, establishing a realistic benchmark to show the substantial improvement of GPP over the enterprise's existing planning system provided by a market-leading vendor. In counterfactual simulations, GPP plans notably surpass the historical performance of the traditional plans.

This paper is organized as follows: we contrast our work with related research, describe the problem setup, present our methods involving GNN, Offline RL, and Policy Simulations, showcase our experiment results, and conclude.

## Related Work

The representation power of GNN embedding for graph-structured data has resulted in successful applications across a wide range of use cases, including traffic forecasting (Cui et al., 2020), crop yield prediction (Fan et al., 2022), Covid-19 case prediction (Fritz et al., 2022; Kapoor et al., 2020), social recommendation (Song et al., 2019), drug discovery (Gaudelet et al., 2021), biological networks (Zhang et al., 2021), and material science and chemistry (Reiser et al., 2022). In the context of supply chains, supply decisions for each edge should not simply depend on the supply-demand imbalances of the associated source and destination nodes alone but on the states of other influential nodes over the network, considering the dynamics among competing and cooperative nodes within the topological network structures. Individual node-level supply decisions may result in locally optimal outcomes. In GNN, additional graph convolution layers gather messages from distant neighborhood hops, incorporating them into the node's embedding vector (or learned feature vector) through aggregation and transformation (Figure 2). Our primary goal of using GNN for supply chain networks is to effectively represent similar patterns and interactions of supply-demand imbalance states across nodes in analogous topological subgraphs/graphs, all of which are mapped into a close embedding space.

---

[1] Defined in this paper as the propensity of accepting excess-stock risk to mitigate out-of-stock risk, or vice versa.

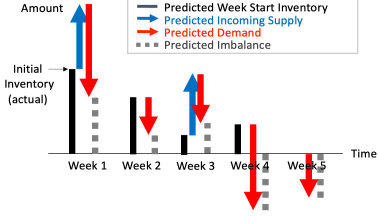

Figure 3: An Example of Predicted Imbalance Over Time.

RL has demonstrated remarkable success in surpassing human-level performance in various gaming contexts (Mnih et al., 2015; D. Silver et al., 2016; Vinyals et al., 2019). Yet, its applications for real-world complex supply chain networks remain relatively limited. Past studies in RL for supply chains have mainly concerned small-scale multi-echelon supply chain optimization for serial or simple diverging supply chains (Gijsbrechts et al., 2022; Hubbs et al., 2020; Kegenbekov and Jackson, 2021; Khirwar et al., 2023; Oroojlooyjadid et al., 2022; Wu et al., 2023) and local inventory optimization for an individual node's replenishment policy (Demizu et al., 2023; Madeka et al., 2022), overlooking complex SKU-specific and time-varying topological networks where each node has diverse numbers of incoming and outgoing connections. Previous RL studies have also employed explicit state representations based on observable and known features in the chains. However, for large-scale supply chain networks, an explicit state representation may prove inefficient and fail to encompass the complete and nuanced interactions between all nodes. In contrast, GPP utilizes a latent state representation via graph node embedding, achieving a comprehensive representation of node interactions within the network. Moreover, most of the previous research has used Online RL simulators that are considerably simpler than real-world supply chain networks with intricate topological structures, uncertainties, and constraints. Thus, applying Online RL training with simple simulation environments to complex supply chain networks poses crucial challenges. We leverage Offline RL as a remarkably practical approach to train resilient policy models, utilizing historical data that captures uncertainties and intricacies within real-world supply chain networks. The trained GPP can be instantly applied to generate globally optimal supply plans in real-time for any probabilistically changing states and conditions in the network over time. This type of swift, scalable, and dynamic planning with trained GPP policies is in sharp contrast to conventional OR-based supply planning methods (MIP/CP), which solve for each new network state as a fresh optimization problem, resulting in inefficient and unscalable computations even for deterministic constraints and cost optimizations.

## Methods

### Problem Description

We consider supply chain networks where each SKU is associated with its own distinct topological graph. Let $\mathcal{G} = (\mathcal{V}, \mathcal{E})$ denote a directional graph for the SKU-specific supply chain network,[2] containing a collection of nodes $\mathcal{V}=\{1,...,n\}$ with diverse node types (e.g., plants, distribution centers, retailers) and edges $\mathcal{E} \in \mathcal{V} \times \mathcal{V}$ where $(v,w) \in \mathcal{E}$ denotes an edge from a source node $v$ to a destination node $w$. The nodes and edges in $\mathcal{G}$ at time $t$ are associated with a set of node feature vectors $\boldsymbol{x}^t = \{x_v^t \in \mathbb{R}^K : v \in \mathcal{V}\} \in \mathbb{R}^{K \times |\mathcal{V}|}$ and a set of edge feature vectors $\boldsymbol{a}^t = \{a_{vw}^{\lambda,t} \in \mathbb{R}^M : (v,w) \in \mathcal{E}, \lambda \in \Lambda\} \in \mathbb{R}^{M \times |\Lambda| \times |\mathcal{E}|}$ for different risk preferences $\lambda \in \Lambda = \{1,2,...,|\Lambda|\}$ (associated with parameterized node-level reward functions, as illustrated in Figure 4) where $K$ is the number of features to define a node feature vector and $M$ is the number of distinct modes of transportation (MOT). Each risk preference setting $\lambda$ is associated with specific reward functions involving different cost weights for out-of-stock (lost sales) and excess-stock outcomes. The edge feature vector $a_{vw}^{\lambda,t} \in \mathbb{R}^M$ has the outgoing supply amounts of each MOT for an edge from node $v$ to node $w$ at time $t$, affected by risk preference $\lambda$. That is, $a_{vw}^{\lambda,t}[m] \in \mathbb{R}$ is the action corresponding to an MOT, $m \in \{1,2,...,M\}$. Also, $\boldsymbol{a}^t$ indicates outgoing supply amounts of each MOT across all the edges and risk preferences, which is viewed as the *actions* (decision variables) over the network in our RL policy. In addition, let $\mathcal{G}^R = (\mathcal{V}, \mathcal{E}^R)$ denote the reverse graph of $\mathcal{G}$ with the same nodes and features but with the directions of all edges reversed, that is, $\mathcal{E}^R = \{(w,v) | (v,w) \in \mathcal{E}\}$ and a set of edge features $\boldsymbol{b}^t = \{b_{wv}^{\lambda,t} \in \mathbb{R}^M : b_{wv}^{\lambda,t} = a_{vw}^{\lambda,t}$ for $(w,v) \in \mathcal{E}^R, \lambda \in \Lambda\}$.

Node feature vector $x_v^t = \left(f_v^{t|t}, f_v^{t+1|t}, ..., f_v^{t+K-1|t}\right) \in \mathbb{R}^K$ for node $v$ is defined in *predicted imbalance features* $f_v^{t+k|t} \in \mathbb{R}$ for $k = 0, 1, ..., K-1$, representing a forward-looking temporal profile of supply-demand imbalance at node $v$, spanning from time $t$ to $t+K-1$, as predicted at time $t$ before taking action $\boldsymbol{a}^t$ (Figure 3). That is, $f_v^{t|t} = I_v^t = $ initial inventory at node $v$ at the start of interval $t$, and $f_v^{t+k|t} = f_v^{t+k-1|t} + S_v^{t+k-1|t} - D_v^{t+k-1}$ for $k = 1, 2,..., K-1$, where $D_v^{t+k-1}$ is the demand at time $t+k-1$ and $S_v^{t+k-1|t}$ is the incoming supply which was sent from source nodes before $t$ and received at node $v$ at time $t+k-1$ with some lead time. Specifically,[3] for any received time $j$ ($\geq t-1$), $S_v^{j|t} = \sum_{w \in \mathcal{N}_{\text{src}}(v)} \sum_{m=1}^{M} \sum_{t'=0}^{t-1} a_{wv}^{t'}[m] \delta\{\tau_{wv}^{t'}[m] = j - t'\}$ where $\tau_{wv}^{t'}[m]$ is the lead time of MOT $m$ for outgoing supply at time $t'$ from all neighboring source nodes $w \in \mathcal{N}_{\text{src}}(v)$ to node $v$, and unit sample function $\delta\{n\} = 1$ if $n$ is true; otherwise, 0.

---

[2] For the ease of notation and without loss of generality, we drop the SKU-specific subscript from most notations, unless we explicitly include it.

[3] For the ease of notation and without loss of generality, we drop the risk preference $\lambda$ notation from some math expressions including actions.

Predicted imbalance features are not only influenced by the initial inventory, but also by the predicted future demands and the predicted future incoming supplies that are currently in-transit. For example, if the predicted demands are significantly high (or low), the predicted imbalance features are more likely to be an out-of-stock (or excessive stock) state, respectively. The predicted imbalance features, contingent on demand and other conditions, play a crucial role in directing supply actions throughout the network to achieve a more balanced future network state.

We define the GNN-based RL policy (actor) network $\mu$ as a continuous deterministic function which recommends supply actions, i.e., outgoing supply amounts on each edge $\boldsymbol{a} = \mu(\boldsymbol{x}) \in \mathbb{R}^{M \times |\mathcal{E}| \times |\Lambda|}$ affected by each risk preference $\lambda \in \Lambda = \{1,2,\ldots,|\Lambda|\}$ at a network state $\boldsymbol{x}$ (i.e., predicted imbalances of all nodes). The GNN-based RL value (critic) network $Q(\boldsymbol{x}, \boldsymbol{a}) \in \mathbb{R}^{|\Lambda|}$ predicts the *network-level* expected returns (= the expected cumulative discounted future rewards) for each risk preference $\lambda \in \Lambda$, after taking a supply action $\boldsymbol{a}$ at a network state $\boldsymbol{x}$ and thereafter following policy $\mu$. Our paper is framed within the RL framework where the policy network $\mu$ is trained to recommend actions that optimize the critic network values $Q$ across all possible states.

**Graph Neural Networks (GNN)**
We use Graph Attention Networks (GAT), a type of GNN architecture designed to assign dynamic attention weights to neighboring nodes in graph convolutions. In GAT, provided the node features $\boldsymbol{h}^{(0)} (= \boldsymbol{x})$ for $\mathcal{G} = (\mathcal{V}, \mathcal{E})$ with edges from neighboring source nodes $j \in \mathcal{N}_{\text{src}}(i) = \{j | (j,i) \in \mathcal{E}\}$ to node $i$, $\text{GAT}_X$ denote the $L$-layered embedding network that makes iterative updates $l = 1, 2, \ldots, L$ and calculates $\boldsymbol{h}^{(L)} =$
$$\text{GAT}_X(\boldsymbol{h}^{(0)}, \mathcal{G}\,; \phi_X) = \left\{h_v^{(L)} \in \mathbb{R}^H : v \in \mathcal{V}\right\} \in \mathbb{R}^{H \times |\mathcal{V}|} \quad (1)$$
where $\phi_X$ is the set of all learned parameters of $\text{GAT}_X$ and $H$ is the embedding dimension. Now, the graph node embedding $\boldsymbol{s}$ for node features $\boldsymbol{x}$ is defined as: with concatenation operator $\|$,
$$\boldsymbol{s} \triangleq \text{emb}_X(\boldsymbol{x}; \phi_{X_f}, \phi_{X_b}) = [\boldsymbol{s}_f \| \boldsymbol{s}_b] \in \mathbb{R}^{2H \times |\mathcal{V}|} \quad (1)$$
where $\boldsymbol{s}_f = \text{GAT}_X(\boldsymbol{x}, \mathcal{G}\,; \phi_{X_f})$ and $\boldsymbol{s}_b = \text{GAT}_X(\boldsymbol{x}, \mathcal{G}^R\,; \phi_{X_b})$. $\boldsymbol{s}_f$ and $\boldsymbol{s}_b$ are calculated using the forward-directional (original) graph and the backward-directional (reverse) graph, respectively. The $\boldsymbol{s}_f$ embedding vectors from graph $\mathcal{G}$ incorporate the competing and collaborative dynamics of multiple source nodes to a destination node into their own attentions, whereas the $\boldsymbol{s}_b$ embedding vectors from reversed graph $\mathcal{G}^R$ encompass the interactions of multiple destination nodes to a source node into their respective attentions.

Based on the graph node embedding $\boldsymbol{s} = \text{emb}_X(\boldsymbol{x}) = \{s_v \in \mathbb{R}^{2H} : s_v = [s_{f,v} \| s_{b,v}], v \in \mathcal{V}\}$, we first calculate

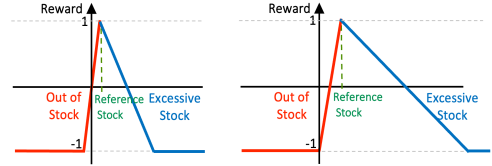

Figure 4: Node-level Reward Functions (Examples)
Left ($\lambda = 1$): $c_1^\lambda = 2, c_2^\lambda = 5, f_{ref}^\lambda = 0.2$, Right ($\lambda = 2$): $c_1^\lambda = 1, c_2^\lambda = 5, f_{ref}^\lambda = 0.4$

unconstrained action $\hat{a}_{vw} = \{\hat{a}_{vw}^\lambda \in \mathbb{R}^M : \lambda \in \Lambda\} = \text{mlp}_\mu([s_v \| s_w]) \in \mathbb{R}^{M \times |\Lambda|}$ from node $v$ to node $w$, where $\text{mlp}_\mu$ is a multilayer feedforward network with sigmoid outputs. Second, for each node $v$ and risk preference $\lambda$, we calculate the sum of all unconstrained outgoing supply amounts from node $v$ to all neighboring destination nodes $w \in \mathcal{N}_{\text{dst}}(v)$ over all MOTs, $A_v^\lambda = \sum_{w \in \mathcal{N}_{\text{dst}}(v)} \sum_{m=1}^M \hat{a}_{vw}^\lambda[m]$. Third, on each time, if $A_v^\lambda$ is greater than the supply capability $Y_v$ ($\triangleq$ inventory $-$ demand) of node $v$ at the time, we should normalize $\hat{a}_{vw}^\lambda$ by $Y_v$ because the total outgoing supply amount from a node should not excess its supply capability. That is, if $A_v^\lambda > 1.0$, $a_{vw}^\lambda = (Y_v / A_v^\lambda) \hat{a}_{vw}^\lambda$; otherwise, $a_{vw}^\lambda = Y_v \hat{a}_{vw}^\lambda$.[4] Now, the policy $\mu$ is defined using $a_{vw}^\lambda$:
$$\boldsymbol{a} = \mu(\boldsymbol{x}) \triangleq \mu_s(\boldsymbol{s}) = \{a_{vw}^\lambda : (v,w) \in \mathcal{E}, \lambda \in \Lambda\} \in \mathbb{R}^{M \times |\mathcal{E}| \times |\Lambda|} \quad (2)$$

In like manner, provided both node features $\boldsymbol{h}^{(0)} (= \boldsymbol{x})$ and edge features $\boldsymbol{e} (= \boldsymbol{a})$ for $\mathcal{G} = (\mathcal{V}, \mathcal{E})$, $\text{GAT}_{XA}$ denotes the $L$-layered embedding network that makes iterative updates $l = 1, 2, \ldots, L$ and calculates $\boldsymbol{h}^{(L)} =$
$$\text{GAT}_{XA}(\boldsymbol{h}^{(0)}, \boldsymbol{e}, \mathcal{G}\,; \phi_{XA}) = \{h_v^{(L)} \in \mathbb{R}^H : v \in \mathcal{V}\} \in \mathbb{R}^{H \times |\mathcal{V}|} \quad (4)$$
where $\phi_{XA}$ is the set of all learned parameters of $\text{GAT}_{XA}$.[5] Then, graph node embedding $\boldsymbol{u}$ for a given pair of node features $\boldsymbol{x}$ and edge features $\boldsymbol{a}$ (supply actions) is defined:
$$\boldsymbol{u} \triangleq \text{emb}_{XA}(\boldsymbol{x}, \boldsymbol{a}; \phi_{XA_f}, \phi_{XA_b}) = [\boldsymbol{u}_f \| \boldsymbol{u}_b] \in \mathbb{R}^{2H \times |\mathcal{V}|} \quad (3)$$
$\boldsymbol{u}_f = \text{GAT}_{XA}(\boldsymbol{x}, \boldsymbol{a}, \mathcal{G}\,; \phi_{XA_f})$, $\boldsymbol{u}_b = \text{GAT}_{XA}(\boldsymbol{x}, \boldsymbol{a}^T, \mathcal{G}^R\,; \phi_{XA_b})$. $\boldsymbol{u}_f$ and $\boldsymbol{u}_b$ each are calculated using the forward-directional (original) and backward-directional (reverse) graphs.

For $\boldsymbol{u} = \text{emb}_{XA}(\boldsymbol{x}, \boldsymbol{a}) = \{u_v \in \mathbb{R}^{2H} : u_v = [u_{f,v} \| u_{b,v}], v \in \mathcal{V}\}$, $q_v = \{q_v^\lambda \in \mathbb{R}: \lambda \in \Lambda\} = \text{mlp}_Q(u_v) \in \mathbb{R}^{|\Lambda|}$ is the node-level returns for $v \in \mathcal{V}$ and all $\lambda \in \Lambda$, and $\text{mlp}_Q$ is a multilayer feedforward network with outputs of tanh multiplied by $1/(1-\gamma)$ where $\gamma$ is the discounting factor in RL. Then, network-level expected returns are:
$$q = Q(\boldsymbol{x}, \boldsymbol{a}) \triangleq Q_u(\boldsymbol{u}) = \{q^\lambda = \sum_{v \in \mathcal{V}} q_v^\lambda : \lambda \in \Lambda\} \in \mathbb{R}^{|\Lambda|} \quad (4)$$

**Offline Reinforcement Learning (Offline RL)**
We use Offline RL to iteratively optimize the GNN-based value network $Q(\boldsymbol{x}, \boldsymbol{a})$ and policy network $\mu(\boldsymbol{x})$ to learn from a historical dataset of network transition samples. Our approach is presented within the DDPG (Lillicrap et al., 2015) framework for continuous action control, open to potential extensions with off-policy algorithms like TD3 (Fujimoto et al., 2018) and SAC (Haarnoja et al., 2018). In this paper we illustrate a simple network-level reward function $r(\boldsymbol{x}^t, \boldsymbol{a}^t; \lambda)$ which relies on the outcome state [6] $\boldsymbol{x}^{t+1}$ alone. We define $r(\boldsymbol{x}^t, \boldsymbol{a}^t; \lambda) = \sum_{v \in \mathcal{V}} \Upsilon(f_v^{t+K|t+1}; \lambda)$ where $\Upsilon$ is a node-level reward function (Figure 4) and $f_v^{t+K|t+1}$ is the $(K-1)$ timestep-ahead imbalance feature of $\boldsymbol{x}^{t+1}$ whose value is influenced by the outcome of current action $\boldsymbol{a}^t$ for any lead time up to $K$ timesteps. The $(c_1^\lambda, c_2^\lambda, f_{\text{ref}}^\lambda)$-parameterized node-level reward function $\Upsilon(.\,;\lambda)$ determines the risk preference of the trained policy, penalizing the excess-stock and out-of-stock outcomes with different costs[7]: for any $f = f_v^{t+K|t+1}$, $\Upsilon(f) = \max\{1 - c_1^\lambda(f - f_{\text{ref}}^\lambda), -1\}$ if $f - f_{\text{ref}}^\lambda \geq 0$ ; otherwise, $\Upsilon(f) = \max\{1 - c_2^\lambda(f_{\text{ref}}^\lambda - f), -1\}$ where $c_1^\lambda (>0)$ and $c_2^\lambda (>0)$ each are excess-stock and out-of-stock cost hyperparameters and $f_{\text{ref}}^\lambda (\geq 0)$ is the reference stock level hyperparameter for risk preference $\lambda$. Thus, we assign the highest node-level reward ($= 1$) to a node if action $\boldsymbol{a}^t$ makes the node's predicted imbalance to the reference stock level (e.g., safety stock). Also, by increasing either the $c_2^\lambda/c_1^\lambda$ ratio or the $f_{\text{ref}}^\lambda$ value, we aim to train a policy that prioritizes avoiding out-of-stock risk over excess stock risk.

---

[4] We also consider the shipping capacity constraint of each MOT for a node in a similar way but with the summed amount per MOT. We may further quantize $a_{vw}$ for SKU-specific minimum order quantity (MOQ) constraint.

[5] $\phi_X$ and $\phi_{XA}$ are distinct sets of parameters, although the same variable notations are employed for simplicity in illustrations. More details on GATs are available in the *Technical Appendix*.

The return from a state $\boldsymbol{x}^t$ at time $t$ for risk preference $\lambda$ is defined as the cumulative discounted future reward $R_t^\lambda = \sum_{i=t}^\infty \gamma^{i-t} r(\boldsymbol{x}^t, \boldsymbol{a}^t; \lambda)$ for a discounting factor $\gamma \in [0,1)$. For a continuous deterministic policy network $\mu$ and the offline training dataset $\mathfrak{D}$,

$$Q(\boldsymbol{x}, \boldsymbol{a}) = \mathbb{E}_{\boldsymbol{x}' \sim P(\cdot|\boldsymbol{x},\boldsymbol{a}), \boldsymbol{a}' \sim \mu(\boldsymbol{x}')}[R_t | \boldsymbol{x}^t = \boldsymbol{x}, \boldsymbol{a}^t = \boldsymbol{a}] \quad (5)$$
$$\approx \mathbb{E}_{(\boldsymbol{x},\boldsymbol{a},\boldsymbol{x}') \sim \mathfrak{D}}[r(\boldsymbol{x},\boldsymbol{a}) + \gamma Q(\boldsymbol{x}', \mu(\boldsymbol{x}'))]$$

where $R_t = \{R_t^\lambda : \lambda \in \Lambda\} \in \mathbb{R}^{|\Lambda|}$, $r(\boldsymbol{x},\boldsymbol{a}) = \{r^\lambda(\boldsymbol{x},\boldsymbol{a}; \lambda) : \lambda \in \Lambda\}$. Here, $r^\lambda(\boldsymbol{x}, \boldsymbol{a}; \lambda)$ is the network-level reward function shaped for risk preference $\lambda$.

Let $\theta_Q = \{\phi_{XA_f}, \phi_{XA_b}, \phi_{\text{mlp}_Q}\}$ and $\theta_\mu = \{\phi_{X_f}, \phi_{X_b}, \phi_{\text{mlp}_\mu}\}$ be the sets of parameters of the value network $Q(\boldsymbol{x}, \boldsymbol{a}; \theta_Q)$ and the policy network $\mu(\boldsymbol{x}; \theta_\mu)$.

We optimize $\theta_Q$ by minimizing the loss $L(\theta_Q)$ of temporal difference (TD) errors over $\mathfrak{D}$ through backprop:

$$L(\theta_Q) = \mathbb{E}_{(\boldsymbol{x},\boldsymbol{a},\boldsymbol{x}') \sim \mathfrak{D}}[|\Lambda|^{-1} \| y - Q(\boldsymbol{x}, \boldsymbol{a}; \theta_Q) \|_2^2] \quad (6)$$
$$= \mathbb{E}_{(\boldsymbol{x},\boldsymbol{a},\boldsymbol{x}') \sim \mathfrak{D}}[|\Lambda|^{-1} \sum_\lambda (y^\lambda - Q^\lambda(\boldsymbol{x}, \boldsymbol{a}; \theta_Q))^2]$$

where $y = r(\boldsymbol{x},\boldsymbol{a}) + \gamma Q(\boldsymbol{x}', \mu(\boldsymbol{x}'; \theta_\mu^{\text{target}}); \theta_Q^{\text{target}})$ and $y^\lambda = r^\lambda(\boldsymbol{x},\boldsymbol{a}; \lambda) + \gamma Q^\lambda(\boldsymbol{x}', \mu^\lambda(\boldsymbol{x}'; \theta_\mu^{\text{target}}); \theta_Q^{\text{target}})$. $y = \{y^\lambda : \lambda \in \Lambda\}$ is the prediction target vector over risk preferences as the sum of the immediate reward vector and the discounted value prediction vector at the next time. Employing $\theta_\mu^{\text{target}}$ and $\theta_Q^{\text{target}}$ (frozen target network parameters) in preparing $y$ samples is necessary for stabilizing optimization. We reset the target network parameters to the current network parameters every fixed number of learning updates. Importantly, we first train the initial $\theta_Q$ using $y = r(\boldsymbol{x},\boldsymbol{a}) + \gamma Q(\boldsymbol{x}', \boldsymbol{a}')$ prepared with historical transition samples $(\boldsymbol{x}, \boldsymbol{a}, \boldsymbol{x}')$ in $\mathfrak{D}$, instead of involving a pre-mature policy network $\mu$. This approach of using the behavioral, or data-generating, policy to train the initial $Q$ turns out to be effective in Offline RL, which is also consistent with the previous work (Brandfonbrener et al., 2021; Goo and Niekum, 2022).

One of the desirable characteristics of the learned policy behavior is that the total supply action across all edges from source nodes $w \in \mathcal{N}_{\text{src}}(v)$ at time $t$, $\sum_{w \in \mathcal{N}_{\text{src}}(v)} \sum_{m=1}^M a_{wv}^{\lambda,t}[m]$ is expected to be zero if the predicted imbalance of node $v$, $f_v^{t+K-1|t} \geq f_{\text{ref}}^\lambda$. If $f_v^{t+K-1|t} < f_{\text{ref}}^\lambda$, the total supply action is expected to compensate for the negative imbalance $f_v^{t+K-1|t} - f_{\text{ref}}^\lambda$. To obtain this desirable behavior, we add a behavioral regularization term[8] $\mathcal{L}^\lambda$ to $Q^\lambda(\boldsymbol{x}, \boldsymbol{a})$:
$\mathcal{L}^\lambda = -\frac{1}{|\mathcal{V}|} \sum_{v \in \mathcal{V}} (\min(f_v^{t+K-1|t} - f_{\text{ref}}^\lambda, 0) + \sum_{w \in \mathcal{N}_{\text{src}}(v)} \sum_{m=1}^M a_{wv}^{\lambda,t}[m])^2$,
$Q^\lambda(\boldsymbol{x},\boldsymbol{a}) \leftarrow Q^\lambda(\boldsymbol{x}, \boldsymbol{a}) + \eta \mathcal{L}^\lambda$ where $\eta$ is a hyperparameter.

We train $\mu^\lambda(\boldsymbol{x}; \theta_\mu) = \arg\max_{\boldsymbol{a}} Q^\lambda(\boldsymbol{x}, \boldsymbol{a})$ by maximizing the expected return from the start state distribution in $\mathfrak{D}$, $J(\theta_\mu) = \mathbb{E}_{\boldsymbol{x} \sim \mathfrak{D}}[\sum_\lambda Q^\lambda(\boldsymbol{x}, \mu^\lambda(\boldsymbol{x}; \theta_\mu))]$ with respect to $\theta_\mu$. That is, the policy parameters are updated along the direction of the $\nabla_{\theta_\mu} J(\theta_\mu)$ via backpropagation:

$$\nabla_{\theta_\mu} J(\theta_\mu) \approx \mathbb{E}_{\boldsymbol{x} \sim \mathfrak{D}}\left[|\Lambda|^{-1} \sum_\lambda \nabla_{\theta_\mu} Q^\lambda(\boldsymbol{x}, \mu^\lambda(\boldsymbol{x}; \theta_\mu))\right] \quad (7)$$
$$= \mathbb{E}_{\boldsymbol{x} \sim \mathfrak{D}}\left[|\Lambda|^{-1} \sum_\lambda \nabla_{\theta_\mu} \mu^\lambda(\boldsymbol{x}; \theta_\mu) \nabla_{\boldsymbol{a}} Q^\lambda(\boldsymbol{x}, \boldsymbol{a})|_{\boldsymbol{a} = \mu^\lambda(\boldsymbol{x}; \theta_\mu)}\right]$$

GAT embedding layers and MLP layers of RL value and policy networks are concurrently trained through end-to-end learning and optimization. Thus, the model learns the optimal graph embedding representations with node and action features in the light of the optimization objective of RL. Refer to GPP training algorithm in the *Technical Appendix*.

**Policy Simulations**
Let $I_v^t$ denote the initial inventory at node $v$ at time interval $t$. Suppose that we are provided with a node-level probabilistic demand prediction model for sampling predicted demands over the time horizon at the start of interval $t$, which are denoted as $\widehat{D}_v^{j|t} \sim p(D_v^{j|t})$. Also, we assume that there is an edge MOT-level probabilistic lead time prediction model for sampling the predicted led times with which we predict the incoming supplies over the time horizon, $\widehat{S}_v^{j|t} \sim p(S_v^{j|t})$, at the start of interval $t$. The incoming supplies are sent from source nodes before $t$ and received at node $v$ with certain lead time.[9]

---

[6] The definition can be extended to rely on the complete network snapshot transition $(\boldsymbol{x}^t, \boldsymbol{a}^t, \boldsymbol{x}^{t+1})$ to include actual transportation costs, etc.
[7] Given the dynamic nature of economic costs for unit out-of-stock and unit excess stock, our strategy entails pre-training the policies with diverse risk parameters. This enables us to select the most suitable policy through simulations, seamlessly adapting to changing objectives.
[8] This regularization term assists the policy in learning optimal behaviors for states with feature values that have sparse historical samples.
[9] Probabilistic demand and lead time prediction models are not in the scope of this paper. We assume the models are separately trained and provided.

*Planning Simulation*: At the start of interval $t$, we generate a supply action plan (i.e., outgoing supplies) over the time horizon ($j = t, ..., t + J - 1$) using the trained policy in iterations. In iteration $j$, at distribution node $v$, we calculate $x_v^j = \left(f_v^{j|j}, f_v^{j+1|j}, ..., f_v^{j+K-1|j}\right)$ using $f_v^{j|j} = \hat{I}_v^j$ = predicted initial inventory at node $v$ in interval $j$, $\widehat{D}_v^{j+k-1|t}$ = predicted demand sample for interval $j + k - 1$ as predicted at the start of interval $t$, $\hat{S}_v^{j+k-1|j}$ = predicted incoming supply sample for interval $j + k - 1$ which was sent from source nodes before interval $j$, where $f_v^{j+k|j} = f_v^{j+k-1|j} + \hat{S}_v^{j+k-1|j} - \widehat{D}_v^{j+k-1|t}$ for $k = 1, ..., K - 1$, and $\hat{I}_v^t = I_v^t$. Then, $a^j = \mu(x^j)$ where $x^j = \{x_v^j \in \mathbb{R}^K : v \in \mathcal{V}\}$. The simulated inventory is updated: $\hat{I}_v^{j+1} = \hat{I}_v^j + \hat{S}_v^{j|j} - \widehat{D}_v^{j|t} - A_v^j$ where $A_v^j = \sum_{w \in \mathcal{N}_{dst}(v)} \sum_{m=1}^M a_{vw}^j[m]$ = outgoing supply from node $v$, and $\hat{S}_v^{j+1|j+1} = \hat{S}_v^{j|j} + \hat{s}_v^j$. Note that $\hat{s}_v^j$ is immediate incoming supply at node $v$ in interval $j$ with zero lead time, that is, the supply that are sent from source nodes in interval $j$ and received at node $v$ in the same interval $j$. For plant node $v$, we assume that the predicted or planned production $\widehat{U}_v^{j|t}$ (production quantity at $v$ for interval $j$ when predicted at $t$) is provided, and the future inventory of production node is reconstructed as: $\hat{I}_v^{j+1} = \hat{I}_v^j + \widehat{U}_v^{j|t} - A_v^j$.

*Evaluation Simulation*: We run counterfactual simulations for evaluating a trained policy over a validation or testing dataset. At the start of any interval $t$ in the evaluation period, we plan over the time horizon of $J$ intervals using *predicted* demand. Then we simulate the execution of planned action for current interval $t$ (= 1$^{st}$ interval in the horizon) given *actual* demand $D_v^t$. Thus, the next simulated inventory is updated as $\hat{I}_v^{t+1} = \hat{I}_v^t + \hat{S}_v^{t|t+1} - D_v^t - A_v^t$. The use of *actual* demand $D_v^t$ enables simulating the constrained supply capability limited by actual demand. In the next iteration at the start of interval $t + 1$, we take a fresh round of planning over the following horizon and simulate the planned action for $t + 1$. We repeat iterations till $t + J - 1$. For plant node $v$, using actual production $U_v^t$, the inventory of production node is updated as: $\hat{I}_v^{t+1} = \hat{I}_v^t + U_v^t - A_v^t$. To evaluate policy performance over the time horizon, we define out-of-stock (lost sales) at distribution node $v$ for interval $t$ as $OOS_v^t = \min\{\hat{I}_v^t + \hat{S}_v^{t|t+1} - D_v^t, 0\}$ shown as 0 or negative value, and the excess stock as $ES_v^t = \max\{\hat{I}_v^{t+1}, 0\}$ shown as 0 or positive. The total cost over all SKUs at timestep $t$ is:

$$Cost = \sum_{sku,v} \{c_{es}^{sku} |ES_v^t[sku]| + c_{oos}^{sku} |OOS_v^t[sku]|\} \quad (8)$$

where $c_{es}^{sku}$ and $c_{oos}^{sku}$, respectively, are SKU-specific unit-ES and unit-OOS economic cost parameters. We can set the values of $c_{es}^{sku}$ and $c_{oos}^{sku}$ based on our specific objectives, such as profit maximization or service level maximization.

We use probabilistic Monte-Carlo (MC) to generate the action plan via one or more trained policies, each with varying risk preferences. Even with identical initial inventory, incorporating probabilistic future demands, lead times, and production conditions results in a diverse set of MC sampled states representing the current probabilistic network state. For each policy, we compute actions and costs from each MC state, using the economic costs of the current objective. The optimal policy is chosen based on the lowest averaged cost over all MC states. The final planned actions are determined by averaging sampled actions from the chosen policy. Refer to GPP planning algorithm in the *Technical Appendix*.

## Experiments

We conducted simulations to compare GPP-generated supply actions with historical actions from a global consumer goods company. The GPP policy was trained on 18 months of historical weekly data from March 2021 to August 2022, validated on the subsequent 4 months, and then tested on a 6-month hold-out dataset from 2023. The performance of GPP-generated actions was measured by (a) total out-of-stock, (b) total excess stock, and (c) total economic cost.

*Dataset*: The dataset covers SKUs, constituting 80% of the total demand and shipments. Each SKU-week combination has a unique network graph topology. The network varies in nodes (2 to 20, median 9) and edges (1 to 60, median 20). Two MOTs exist: "truckload" (80% shipment events) with shorter distances and lead times, and "intermodal" (20% shipment events) with longer distances and lead times. The 13-week ahead demand predictions at the SKU/node level show a Median WMAPE ranging from 30% to 50% by predicted timestep, with higher values for later predictions.

*Modeling*: The value and policy models are trained and deployed across all SKUs, each having individual quantity ranges as well as unique topological networks that may vary over time. All relevant variables, such as demand, supply, inventory, and imbalance features, are scaled to a uniform unit using a single SKU-specific scaler based on the maximum inventory quantity during the training data period. The GNN embedding layers, GAT$_X$ and GAT$_{XA}$, were constructed using the PyTorch Geometric (PyG) library (Fey and Lenssen, 2019; Paszke et al., 2019), using GATv2Conv (Brody et al., 2022). Provided the company's typical economic cost objectives, such as $c_{oos}^{sku}/c_{es}^{sku} = 1$ (Situation 1) or $c_{oos}^{sku}/c_{es}^{sku} = 5$ (Situation 2) across SKUs, we trained GPP involving twelve policies ($\lambda = 1, 2, ..., 12$) with different risk preferences: $c_2^\lambda / c_1^\lambda = 1$ or 5, and reference $f_{ref}^\lambda$ ranging from 0.0 to 0.5 in steps of 0.1. Using the validation dataset, we performed the evaluation simulations to select the optimal risk-preferential policy for each cost objective and determine the epoch and hyperparameters resulting in the minimum validation loss, defined as the lowest averaged cost across all cost objectives. We simulated 13-week horizon starting from each SKU/week network snapshot in the dataset, generating 50 probabilistic MC states at every timestep. Also, the production condition in simulations were constrained to match historical productions. The selected optimal policy for Situations 1 and 2, respectively, employed the risk preference ($c_1^{\lambda_{s1}^*}=10$, $c_2^{\lambda_{s1}^*}=10$, $f_{\text{ref}}^{\lambda_{s1}^*} = 0.3$) and ($c_1^{\lambda_{s2}^*}=10$, $c_2^{\lambda_{s2}^*}=10$, $f_{\text{ref}}^{\lambda_{s2}^*} = 0.5$). In Situation 2, ($c_1^{\lambda}=2$ $c_2^{\lambda}=10$, $f_{\text{ref}}^{\lambda}=0.3$) exhibited reasonable performance, though slightly less effective than ($c_1^{\lambda_{s2}^*}=10$, $c_2^{\lambda_{s2}^*}=10$, $f_{\text{ref}}^{\lambda_{s2}^*} = 0.5$). This suggests that a symmetric reward function with a well-chosen reference parameter could also surpass reward functions with $c_2^{\lambda}/c_1^{\lambda} = c_{\text{oos}}^{\text{sku}}/c_{\text{es}}^{\text{sku}}$. More details on our evaluation metrics, dataset, model training and validation are available in the *Technical Appendix*.

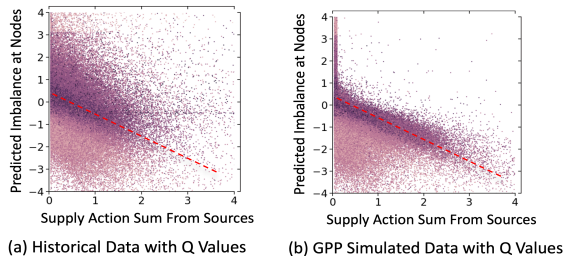

(a) Historical Data with Q Values  (b) GPP Simulated Data with Q Values

Figure 5: Predicted Imbalance at Nodes (y-axis) and Supply Action Sum from Sources (x-axis) with Q Values (color)

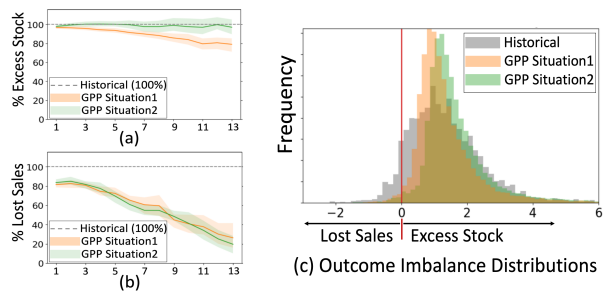

Figure 6: (a) % Excess Stock, (b) % Lost Sales, (c) Outcome Imbalance Distributions (x-axis: scaled imbalance, y-axis: frequency) over SKU/Node/Week Level Data.

*Results*: Using a hold-out testing dataset, we performed the probabilistic evaluation simulations using the selected optimal policy in each situation. Figure 5 depicts trained value and policy network behaviors under the optimal policy for Situation 1, maintaining generality. Figure 5(a) shows the scatter plot of SKU/node/week samples from historical data over the evaluation period. This confirms a successful training of the value network that enables the differentiation of Q values. Darker-colored dots indicate higher Q values, reflecting desirable actions taken, whereas lighter-colored dots represent lower Q values, reflecting undesirable actions taken. The y-axis represents the predicted imbalance states of the sampled nodes, and the x-axis (action in response to the state) represents the sum of outgoing supply action amounts from all source nodes. The diagonal darkest correlated line, with a y-intercept of 0.3 (equal to $f_{\text{ref}}^{\lambda_{s1}^*}$ in the reward function), indicates that the value network successfully learned the ideal collaborative action from source nodes for any predicted imbalance state of the sampled destination node, e.g., addressing a specific negative predicted imbalance (lost sales) is most effectively achieved by an equal aggregated supply. Figure 5(b) shows SKU/node/week samples from GPP simulated data over the same evaluation period. Since GPP policy network was trained to optimize Q values, compared to historical actions in Figure 5(a), GPP actions in Figure 5(b) exhibit a clearly superior alignment with predicted imbalances. Also, GPP effectively suppressed supply to most nodes with positive predicted imbalance (excess stock). These confirm the successful training of the policy network alongside the learned value network.

Figures 6(a) and 6(b) across Situations 1 and 2 illustrate how quickly GPP brings supply-demand balanced network nodes and reductions in excess stock and out-of-stock over timesteps, starting from any historical network state in the evaluation period. In each figure, the baseline of 100% for comparisons represents the historical action performance, calculated as the weekly average of either total excess stock or total lost sales amount across all SKUs and nodes in the network. After the "transient" phase in the initial timesteps, we attain more stable and balanced network states at the 13th timestep: In Situation 1, GPP reduced lost sales by 75% and excess stock by 20%; In Situation 2, GPP reduced lost sales by 81% and excess stock by 4% (all $p$s < .001). Furthermore, Figure 6(c) shows the comparisons between the outcome imbalance distributions of GPP with different risk preferences (Situations 1 and 2) vs. historical actions across SKUs at the 13th timestep. The x-axis shows positive and negative imbalance values (scaled per SKU) for excess stock and lost sales, respectively. The y-axis shows the frequency of SKU/node/week-level samples. The imbalance distributions of GPP actions are much narrower and have thinner tails than that of historical actions. Also, compared to the GPP distribution for Situation 1 ($f_{\text{ref}}^{\lambda_{s1}^*} = 0.3$), the one for Situation 2 ($f_{\text{ref}}^{\lambda_{s2}^*} = 0.5$) is shifted to the right, minimizing lost sales outcomes. This demonstrates GPP's adeptness in selecting optimal policies for each objective situation.

As another point of comparison, we carried out simulations using a rule-based policy where each node calculates SKU/node-specific safety stock based on predicted total demand using the safety days of supply (DOS) information that the company provided. This fixed rule-based policy (not adapting to cost objectives) is considerably deficient, leading to a significant 166% rise in lost sales, despite achieving a 36% reduction in excess stock ($p$s < .001). This highlights the complexity of the underlying supply chain planning challenge. In-depth GPP and rule-based evaluation results with plots are available in the *Technical Appendix*.

## Conclusion

The role of AI in the supply chain planning space has been ambiguous and poorly comprehended. Most current planning systems rely on local rules and OR-based methods that have been in use for decades. GPP, a novel Generative AI technique, is specifically tailored for supply chain planning, which leverages attention-based GNN, Offline RL and policy simulations. Our work plays a pivotal role in shaping the trajectory of AI adoption within the supply chain domain.

# Technical Appendix

**Table of Contents**

# A. Graph Attention Networks

We use Graph Attention Networks (GAT), a type of GNN architecture designed to assign dynamic attention weights to neighboring nodes in graph convolutions. In GAT, provided the node features $\boldsymbol{h}^{(0)}(=\boldsymbol{x})$ for $\mathcal{G}=(\mathcal{V},\mathcal{E})$ with edges from neighboring source nodes $j \in \mathcal{N}_{\text{src}}(i) = \{j|(j,i) \in \mathcal{E}\}$ to node $i$, the update at layer $(l)$ is:

$$h_i^{(l)} = \sigma\left(\alpha_{ii}^{(l-1)}\mathbf{W}_0^{(l-1)}h_i^{(l-1)} + \sum_{j \in \mathcal{N}_{\text{src}}(i)} \alpha_{ij}^{(l-1)}\mathbf{W}_1^{(l-1)}h_j^{(l-1)}\right)$$

where $\alpha_{ij}^{(l-1)} = \text{softmax}_j\left(o_{ij}^{(l-1)}\right) \in \mathbb{R}$ is a dynamic attention[1] for edge $(j,i)$, and

$$o_{ij}^{(l-1)} = \mathbf{c}^{\text{tr}\,(l-1)}\text{LeakyReLU}\left(\mathbf{W}_0^{(l-1)}h_i^{(l-1)} + \mathbf{W}_1^{(l-1)}h_j^{(l-1)}\right).$$

Dynamic attentions enable a node to selectively attend to its neighboring nodes that are highly relevant or predicted to have substantial supply-demand imbalance.

Let $\text{GAT}_X$ denote the $L$-layered embedding network that makes iterative updates $l = 1, 2, \ldots, L$ and calculates

$$\boldsymbol{h}^{(L)} = \text{GAT}_X\left(\boldsymbol{h}^{(0)}, \mathcal{G}\,;\, \phi_X\right) = \left\{h_v^{(L)} \in \mathbb{R}^H : v \in \mathcal{V}\right\} \in \mathbb{R}^{H \times |\mathcal{V}|}$$

where $\phi_X = \{\mathbf{W}_0^{(l-1)}, \mathbf{W}_1^{(l-1)}, \mathbf{c}^{\text{tr}\,(l-1)} : l \in \{1, 2, \ldots, L\}\}$ is the set of all learned parameters of $\text{GAT}_X$ and $H$ is the embedding dimension. Now, the graph node embedding $\boldsymbol{s}$ for node features $\boldsymbol{x}$ is defined as: with concatenation operator $\|$,

$$\boldsymbol{s} \triangleq \text{emb}_X(\boldsymbol{x}; \phi_{X_f}, \phi_{X_b}) = [\boldsymbol{s}_f \| \boldsymbol{s}_b] \in \mathbb{R}^{2H \times |\mathcal{V}|}$$

where $\boldsymbol{s}_f = \text{GAT}_X(\boldsymbol{x}, \mathcal{G}\,;\, \phi_{X_f})$ and $\boldsymbol{s}_b = \text{GAT}_X(\boldsymbol{x}, \mathcal{G}^R\,;\, \phi_{X_b})$.

$\boldsymbol{s}_f$ and $\boldsymbol{s}_b$ are calculated using the forward-directional (original) graph and the backward-directional (reverse) graph, respectively. The $\boldsymbol{s}_f$ embedding vectors from graph $\mathcal{G}$ incorporate the competing and collaborative dynamics of multiple source nodes to a destination node into their own attentions, whereas the $\boldsymbol{s}_b$ embedding vectors from reversed graph $\mathcal{G}^R$ encompass the interactions of multiple destination nodes to a source node into their respective attentions.

In like manner, provided both node features $\boldsymbol{h}^{(0)}(=\boldsymbol{x})$ and edge features $\boldsymbol{e}(=\boldsymbol{a})$ for $\mathcal{G}=(\mathcal{V},\mathcal{E})$, the layer $(l)$ is:

$$h_i^{(l)} = \sigma(\alpha_{ii}^{(l-1)}\mathbf{W}_0^{(l-1)}h_i^{(l-1)} + \sum_{j \in \mathcal{N}_{\text{src}}(i)} \alpha_{ij}^{(l-1)}(\mathbf{W}_1^{(l-1)}h_j^{(l-1)} + \mathbf{W}_2^{(l-1)}e_{ji}))$$

where $\alpha_{ij}^{(l-1)} = \text{softmax}_j\left(o_{ij}^{(l-1)}\right) \in \mathbb{R}$ is a dynamic attention, using

$$o_{ij}^{(l-1)} = \mathbf{c}^{\text{tr}\,(l-1)} \text{LeakyReLU}\left(\mathbf{W}_0^{(l-1)}h_i^{(l-1)} + \mathbf{W}_1^{(l-1)}h_j^{(l-1)} + \mathbf{W}_2^{(l-1)}e_{ji}\right).$$

$\text{GAT}_{XA}$ denotes the $L$-layered embedding network that makes iterative updates $l = 1, 2, \ldots, L$ and calculates

$$\boldsymbol{h}^{(L)} = \text{GAT}_{XA}\left(\boldsymbol{h}^{(0)}, \boldsymbol{e}, \mathcal{G}\,;\, \phi_{XA}\right) = \{h_v^{(L)} \in \mathbb{R}^H : v \in \mathcal{V}\} \in \mathbb{R}^{H \times |\mathcal{V}|}$$

where $\phi_{XA} = \{\mathbf{W}_0^{(l-1)}, \mathbf{W}_1^{(l-1)}, \mathbf{W}_2^{(l-1)}, \mathbf{c}^{\text{tr}\,(l-1)} : l \in \{1, 2, \ldots, L\}\}$ is the set of all learned parameters of $\text{GAT}_{XA}$.[2] Then, the graph node embedding $\boldsymbol{u}$ for a given pair of node features $\boldsymbol{x}$ and edge features $\boldsymbol{a}$ (supply actions) is defined as:

$$\boldsymbol{u} \triangleq \text{emb}_{XA}(\boldsymbol{x}, \boldsymbol{a}; \phi_{XA_f}, \phi_{XA_b}) = [\boldsymbol{u}_f \| \boldsymbol{u}_b] \in \mathbb{R}^{2H \times |\mathcal{V}|}$$

$\boldsymbol{u}_f = \text{GAT}_{XA}(\boldsymbol{x}, \boldsymbol{a}, \mathcal{G}\,;\, \phi_{XA_f})$, $\boldsymbol{u}_b = \text{GAT}_{XA}(\boldsymbol{x}, \boldsymbol{a}^T, \mathcal{G}^R\,;\, \phi_{XA_b})$.

$\boldsymbol{u}_f$ and $\boldsymbol{u}_b$ each are calculated using the forward-directional (original) and backward-directional (reverse) graphs.

---

[1] This can be extended to multi-head attentions, as used in our experiments.
[2] $\phi_X$ and $\phi_{XA}$ are distinct sets of parameters, although the same variable notations are employed for simplicity in illustrations.

# B. Dataset Details

The RL applications for real-world complex supply chain networks remain relatively limited. Past studies in RL for supply chains have mainly concerned small-scale multi-echelon supply chain optimization for serial or simple diverging supply chains (Gijsbrechts et al., 2022; Hubbs et al., 2020; Kegenbekov and Jackson, 2021; Khirwar et al., 2023; Oroojlooyjadid et al., 2022; Wu et al., 2023) and local inventory optimization for an individual node's replenishment policy (Demizu et al., 2023; Madeka et al., 2022), overlooking complex SKU-specific and time-varying topological networks where each node possesses diverse numbers of incoming and outgoing connections. Therefore, our experiments capitalize on historical actual data from a global consumer goods company with complex supply chain networks to demonstrate that GPP accomplishes objective-adapable, probabilistically resilient, and dynamic planning for supply chain networks.

Moreover, most of the previous research used Online RL simulators that were considerably simpler than real-world supply chain networks with complex topological structures, uncertainties, and constraints. Thus, applying Online RL training with simple simulation environments to complex supply chain networks poses crucial challenges. We leverage Offline RL as a remarkably practical approach to train resilient policy models, utilizing historical actual data that captures the uncertainties and intricacies within real-world supply chain networks. To the best of our knowledge, there is no publicly accessible alternative dataset that meets our scientific requirements.

Our dataset contains information on 500 products (SKUs) constituting 80% of the total demand and shipments, with 106 distinct nodes (75 "DC" nodes; 31 "PRODUCTION"). The SKU-specific supply chain network varies in the number of nodes (2 to 20) and the number of edges (1 to 60), as illustrated in Figure 1 and Figure 2. In addition, we exemplify four SKU-specific networks with varying topological complexities from our dataset in Figure 3.

In our models, we allow each SKU-date combination to have a unique network graph topology. We used the stock transfer quantity data (described in Table 1) to define the relevant network of a SKU in a week. If a lane was used in the preceding 13 weeks or is planned to be used in the subsequent 13 weeks, we consider that lane (as an edge) and its associated nodes to be part of the current topology.

Table 1 outlines essential variables within our dataset across various levels of information.

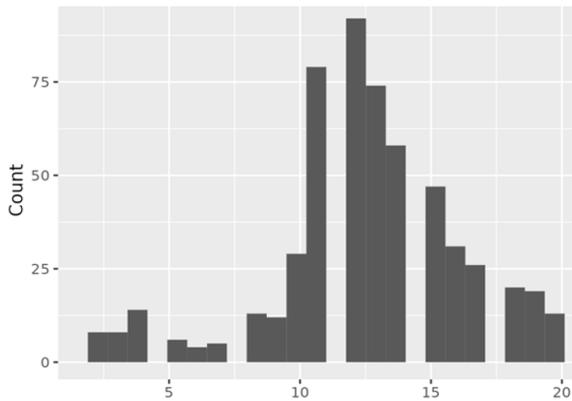

Figure 1: Distribution of Distinct Nodes per SKU.

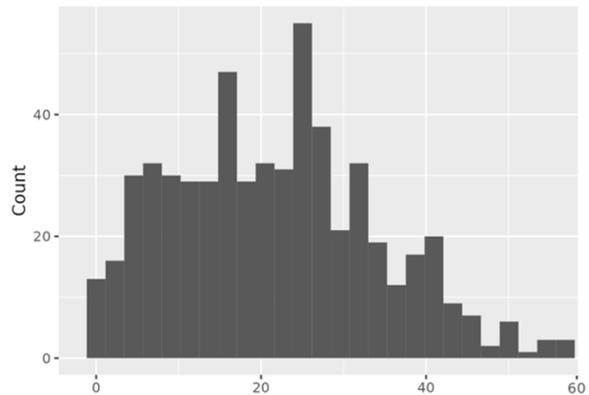

Figure 2: Distribution of Distinct Edges (Source, Destination) per SKU.

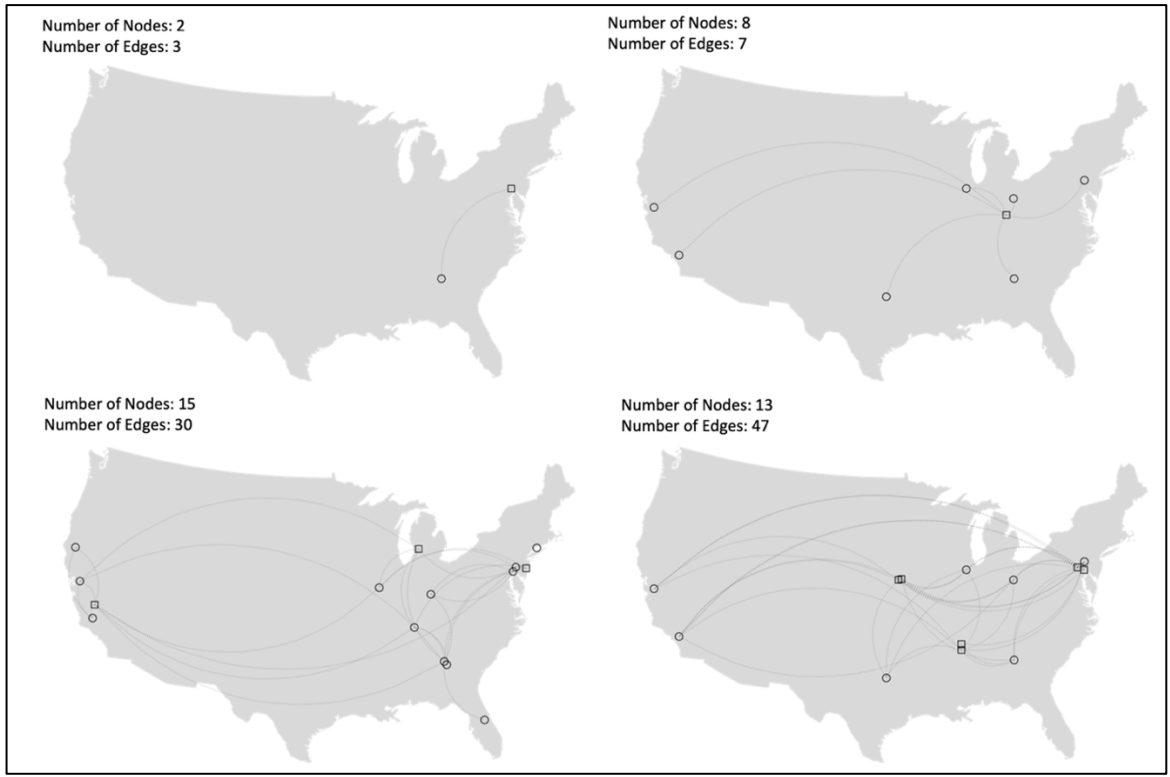

Figure 3: SKU-Specific Networks with Varying Topological Complexities (Examples from the Dataset).

Table 1: The Description of Essential Variables in the Dataset.

| Information Level | Variable Name | Description |
|---|---|---|
| Product | Net price | The average selling price of the product. |
| Node | Node type | A node can be either a distribution center ("DC") or a production node ("PRODUCTION"). |
| Product, node, and date | Inventory quantity | The original data is available for each product and node, at a daily temporal level. We filter the data to get the node inventory at the start of the week (Sunday). Refer to Figure 4. |
| | Demand quantity (Actual demand) | The demand refers to direct customer demand, excluding the demand from other DCs. The original data is available at an individual order level for each product and node, at a daily temporal level. We aggregated the data to the total quantity of orders for the product at the node from all customers, aggregated over the week (Sunday to Saturday). |
| | Demand prediction quantity | The original data has a 13-week-forward horizon from each week of demand predictions. We directly use these. The GPP planning phase uses only the demand prediction quantity, not the actual demand quantity. For SKU/node-level demand prediction errors by predicted timestep, the Median WMAPE varies from 30% to 50%, showing increased values for the later predicted timesteps. Refer to Figure 5. |
| | Production quantity | We aggregated the weekly outgoing stock transfer quantity from production nodes. |

| Product, lane, and date | Stock transfer quantity | The original data is available at an individual order level for each product and lane at a daily temporal level. Each lane is defined by a source node, a destination node, and a mode of transportation (MOT), which can be either "truckload" or "intermodal": "truckload" (80% shipment events) with shorter distances and lead times, and "intermodal" (20% shipment events) with longer distances and lead times. We first aggregated the data to the total quantity of product sent from the source node to the destination node per MOT; and then we aggregated over the week the product shipped from the source node ("ship week;" Sunday to Saturday) and the week the product arrived at the destination node ("delivery week"; Sunday to Saturday). |
|---|---|---|
| | Lead time | For each stock transfer we calculated lead time as the difference between delivery week and ship week. For example, if the delivery week and ship week are the same, lead time is 0; if the delivery week is one week ahead of the ship week, lead time is 1; and so on. During modeling, we create a distribution for each (product, lane), and sample from it to generate a lead time, which we apply to the GPP shipments. Refer to Figure 6. |

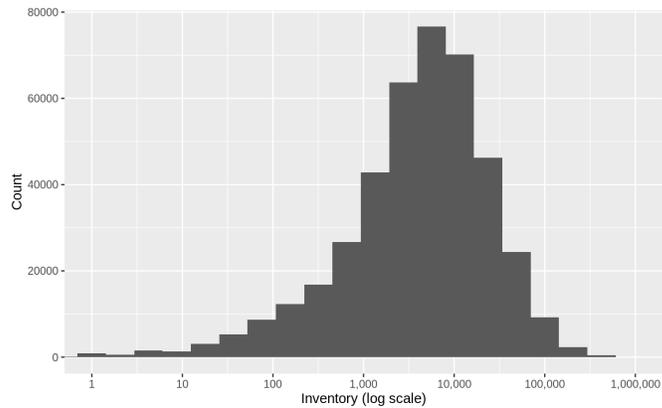

Figure 4: The Inventory Quantity Distribution (SKU/Node/Week-Level Samples)

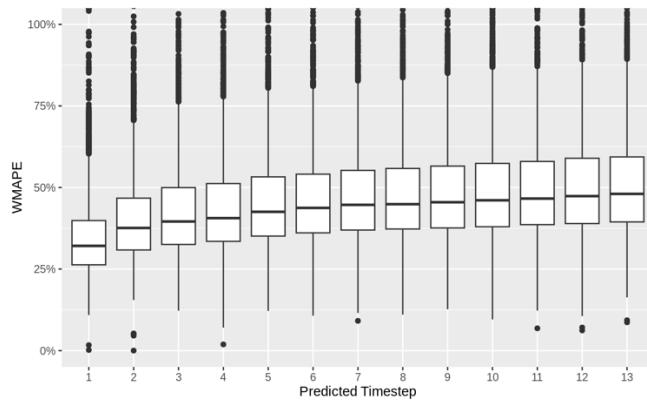

Figure 5: The Demand Prediction WMAPE Distribution (SKU/Node-Level) By Predicted Timestep

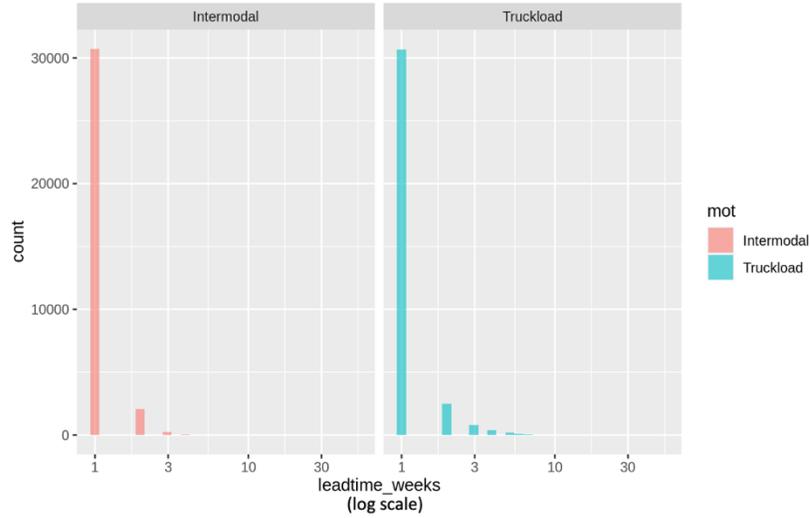

Figure 6: Lead Time Distributions for MOTs (in weeks).

The GPP policy was trained across 500 SKUs on 18 months of historical *weekly* data from March 2021 to August 2022, validated on the subsequent 4 months, and then evaluated on a 6-month hold-out dataset from 2023.

- Training period: Mar 2021 to Aug 2022 (Total 78 Weeks × 500 SKUs = 39,000 Weekly Network Snapshots)

- Validation period: Sep 2022 to Dec 2022 (Total 17 Weeks × 500 SKUs = 8,500 Weekly Network Snapshots)

- Testing period: Jan 2023 to Jun 2023 (Total 26 Weeks × 500 SKUs = 13,000 Weekly Network Snapshots)

The value and policy models are trained and deployed across all SKUs, each having individual quantity ranges as well as unique topological networks that may vary over time. All relevant variables, considered within the imbalance calculation of the nodes such as inventory, demand, and shipped amounts, are scaled to a uniform unit using a single SKU-specific scaler based on the maximum inventory quantity of the SKU during the training data period. In specific, for a given SKU, let $I_{max}^{sku}$ and $I_{min}^{sku}$ denote the maximum and minimum historical inventory quantities across weeks. $I_{min}^{sku} = 0$ for all SKUs. Then, the scaled variables per SKU are defined as: $\tilde{y}^{sku} = y^{sku} / I_{max}^{sku}$.

## C. Model Training and Validation Details

Provided the company's typical economic cost objectives such as $c_{oos}^{sku}/c_{es}^{sku} = 1$ (Situation 1) or $c_{oos}^{sku}/c_{es}^{sku} = 5$ (Situation 2) for all SKUs, we trained GPP involving twelve policies ($\lambda$ =1, 2, ..., 12) with different risk preferences: $c_2^\lambda/c_1^\lambda$ = 1 or 5, and reference $f_{ref}^\lambda$ ranging from 0.0 to 0.5 in steps of 0.1. Figure 7 depicts six out of a total of twelve risk preference configurations in our modeling. The exact risk preference parameters we employed in the model training and validation can be found in **Appendix D**. Note that $c_{es}^{sku}$ and $c_{oos}^{sku}$ are SKU-specific unit-ES (excess stock) and unit-OOS (out-of-stock) economic cost parameters, while $c_1^\lambda$ and $c_2^\lambda$ are risk preference parameters for configuring the reward functions. Our empirical findings suggest that employing the configurations of $c_2^\lambda/c_1^\lambda = c_{oos}^{sku}/c_{es}^{sku}$ along with varying reference $f_{ref}^\lambda$ values is a very effective practice for identifying optimal or nearly optimal policies for a given economic cost objective specified by ($c_{es}^{sku}$, $c_{oos}^{sku}$). Still, exploring the most effective strategy for configuring risk preferences in reward functions for specific economic cost objectives remains a focus of our future research.

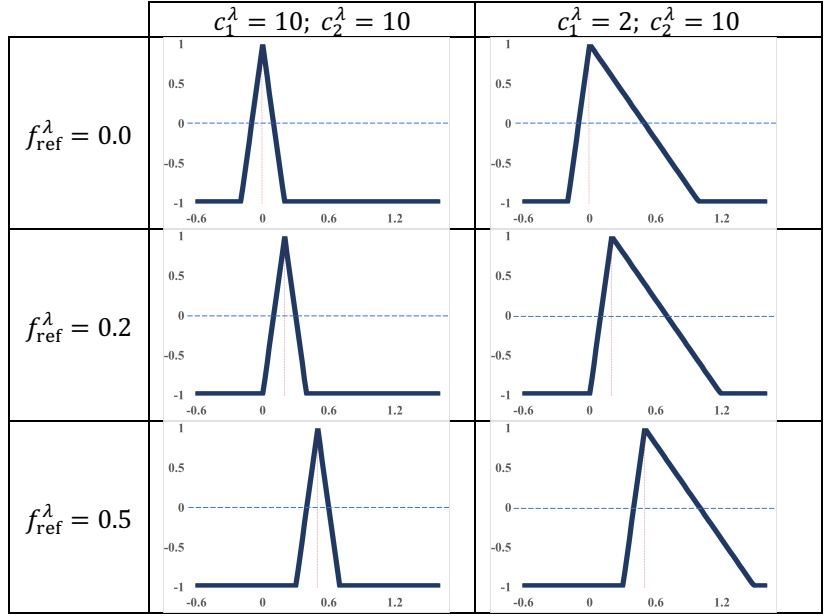

Figure 7: The Reward Functions with Different Risk Preference Configurations (Illustrating only 6 out of the total 12).

We determined the number of predicted imbalance features $K = 4$, considering the lead time distribution whose values tend to be up to 4 weeks.

Setting RL discounting factor $\gamma$ to 0.95, using the validation dataset, we performed probabilistic evaluation simulations over 13-week horizon, starting from each SKU/week network snapshot in the validation period. In simulations, we constrained the production condition to match historical productions. We selected the optimal risk-preferential policy for each cost objective, together with the epoch and hyperparameters resulting in the minimum validation loss, i.e., the lowest averaged value of total costs from the best risk-preferential policy for each economic cost objective at the 13[th] timestep evaluation simulation over the validation period. That is,

$$\text{Validation loss} = \frac{1}{\#\text{objectives}} \frac{1}{T} \sum_{\text{objective}_i} \sum_{t \text{ in validation}} \min_\lambda Cost^{t+12|t}(\lambda \,|\text{objective}_i, t)$$

where $Cost^{t+12|t}(\lambda\,|\text{objective}_i, t)$ is the total cost calculated for the cost objective$_i$ at the 13[th] timestep of the simulation with the $\lambda$-policy (= policy with risk preference $\lambda$), starting from any initial timestep $t$ in the validation period, and $T$ is the total number of initial timesteps (weeks) for the validation.

Using the Adam optimizer, we experimented with learning rates within the range of 1e-4 to 1e-2, incremented in factors of 10, varying the batch sizes among {4, 8, 16, 32} and the behavioral regularization parameter $\eta$ among {0.1, 1, 10}.

$GAT_X$ and $GAT_{XA}$ incorporated LeakyReLU activation applied between each convolution layer. Using 2 or 3 convolution layers with 3 or 8 multi-head attentions for $GAT_X$ and $GAT_{XA}$, we also varied the hidden dimension size. The $mlp_\mu$ for the policy network used 2 or 3 hidden layers with LeakyReLU activation before the output layer with a Sigmoid output function. The $mlp_Q$ for the value network had 2 or 3 hidden layers with LeakyReLU activation before the output layer with a Tanh output function, multiplied by $1/(1-\gamma)$ where $\gamma$ is the discounting factor in RL.

We allowed the model training to run for 100 epochs with early stopping whenever the validation loss plateaued. Importantly, we first train the initial $\theta_Q$ using $y = r(x,a) + \gamma Q(x',a')$ prepared with historical transition samples $(x,a,x')$ in $\mathfrak{D}$, instead of involving a pre-mature policy network $\mu$. We employed the first 10 epochs to train the value network alone, which determines the number of iterations, N in **Appendix E** (GPP Training Algorithm). This approach of using the behavioral, or data-generating, policy to train the initial $Q$ turns out to be effective in Offline RL, which is also consistent with the previous work (Brandfonbrener et al., 2021; Goo and Niekum, 2022).

Through the evaluation simulations on the validation dataset, GPP selects the optimal risk preference for each objective. The selected optimal policy for Situation 1 (S1) employed the risk preference ($c_1^{\lambda_{s1}^*}=10$, $c_2^{\lambda_{s1}^*}=10$, $f_{ref}^{\lambda_{s1}^*}=0.3$), whereas, for Situation 2 (S2), it was with ($c_1^{\lambda_{s2}^*}=10$, $c_2^{\lambda_{s2}^*}=10$, $f_{ref}^{\lambda_{s2}^*}=0.5$). In Situation 2, ($c_1^\lambda=2$, $c_2^\lambda=10$, $f_{ref}^\lambda=0.3$) exhibited reasonable performance, though slightly less effective than ($c_1^{\lambda_{s2}^*}=10$, $c_2^{\lambda_{s2}^*}=10$, $f_{ref}^{\lambda_{s2}^*}=0.5$). This suggests that a symmetric reward function with a well-chosen reference parameter could also surpass reward functions with $c_2^\lambda/c_1^\lambda = c_{oos}^{sku}/c_{es}^{sku}$.

Table 2 summarizes the selected optimal hyperparameter setting.

Table 2: The Best Hyperparameter Setting.

| Hyperparameter name | Selected value |
|---|---|
| discounting factor $\gamma$ | 0.95 |
| soft target update parameter $\tau$ | 0.00005 |
| epochs | 64 |
| learning rate | 0.001 |
| batch size | 4 |
| behavioral regularization parameter $\eta$ | 1 |
| # multi-head attentions | 3 |
| $GAT_X$ graph convolution layers | [16, 16, 16] |
| $GAT_{XA}$ graph convolution layers | [100, 20, 20] |
| $mlp_\mu$ for the policy network | [32, 8] |
| $mlp_Q$ for the value network | [128, 32, 8] |
| optimal risk preference parameters for Situation 1, $\lambda_{s1}^*$ | $c_1^{\lambda_{s1}^*}=10$, $c_2^{\lambda_{s1}^*}=10$, $f_{ref}^{\lambda_{s1}^*}=0.3$ |
| optimal risk preference parameters for Situation 2, $\lambda_{s2}^*$ | $c_1^{\lambda_{s2}^*}=10$, $c_2^{\lambda_{s2}^*}=10$, $f_{ref}^{\lambda_{s2}^*}=0.5$ |

# D. Model Testing and Policy Evaluation Metrics

The company's objectives can shift periodically, seeking to optimize profits or service levels on a quarterly or monthly basis, while considering the trade-offs between minimizing lost sales and managing excess stock risks. A reduction in lost sales leads to higher revenue and margin, while a reduction in excess stocks leads to lower inventory holding costs. We establish a parameterized economic cost objective function that encompasses the costs of lost sales (opportunity cost) and excess stock, with parameters that reflect the company's economic objectives.

Regarding the GPP policy evaluation metrics, to define % metrics relative to historical action performance, we first set the baselines of 100% level to historical weekly sample averages of the total excess stock amount, the total out-of-stock amount, and the total economic cost, aggregated over all SKUs and network nodes. Using *historical* excess stock $ES_{v,\text{hist}}^t[\text{sku}]$ (0 or positive value) and out-of-stock $OOS_{v,\text{hist}}^t[\text{sku}]$ (0 or negative value) from historical actions, $T$ = the total number of weeks in the evaluation period,

100% level Total Excess Stock corresponds to $ES_{\text{hist}} = \frac{1}{T}\sum_{\text{sku},v,t}|ES_{v,\text{hist}}^t[\text{sku}]|$

100% level Total Lost Sales corresponds to $OOS_{\text{hist}} = \frac{1}{T}\sum_{\text{sku},v,t}|OOS_{v,\text{hist}}^t[\text{sku}]|$

100% level Total Cost corresponds to $Cost_{\text{hist}} = \frac{1}{T}\sum_{\text{sku},v,t} \text{price}^{\text{sku}}\{c_{\text{es}}^{\text{sku}}|ES_{v,\text{hist}}^t[\text{sku}]| + c_{\text{oos}}^{\text{sku}}|OOS_{v,\text{hist}}^t[\text{sku}]|\}$
where $\text{price}^{\text{sku}}$ is the unit price of SKU.

The % metrics for GPP at the $j^{th}$ (= 1, 2, …, 13) timestep in the time horizon are calculated as GPP policy's weekly sample averages of the total excess stock amount, the total lost sales amount, and the total economic cost, aggregated over all SKUs and network nodes at the simulation timestep $j$. The averages are based on all simulated weekly states of the networks at the $j^{th}$ timestep, starting from any initial historical network state in the evaluation period. Thus, using $ES_{v,\text{GPP}}^{t+j-1|t}[\text{sku}]$ (0 or positive value) and $OOS_{v,\text{GPP}}^{t+j-1|t}[\text{sku}]$ (0 or negative value), which denote the excess and out-of-stock amounts in node $v$ at the $j^{th}$ timestep following GPP actions, starting from initial timestep $t$, respectively,

% Total Excess Stock of GPP at the $j^{th}$ timestep = $\frac{100}{T} \frac{\sum_{\text{sku},v,t}|ES_{v,\text{GPP}}^{t+j-1|t}[\text{sku}]|}{ES_{\text{hist}}}$

% Total Lost Sales of GPP at the $j^{th}$ timestep = $\frac{100}{T} \frac{\sum_{\text{sku},v,t}|OOS_{v,\text{GPP}}^{t+j-1|t}[\text{sku}]|}{OOS_{\text{hist}}}$

% Total Cost of GPP at the $j^{th}$ timestep = $\frac{100}{T} \frac{\sum_{\text{sku},v,t} \text{price}^{\text{sku}}\{c_{\text{es}}^{\text{sku}}|ES_{v,\text{GPP}}^{t+j-1|t}[\text{sku}]| + c_{\text{oos}}^{\text{sku}}|OOS_{v,\text{GPP}}^{t+j-1|t}[\text{sku}]|\}}{Cost_{\text{hist}}}$
where $\text{price}^{\text{sku}}$ is the unit price of SKU.

For Situation 1 we calculate the cost using the economic cost objective using $c_{\text{oos}}^{\text{sku}}/c_{\text{es}}^{\text{sku}}$ = 1, whereas Situation 2 involves $c_{\text{oos}}^{\text{sku}}/c_{\text{es}}^{\text{sku}}$ = 5 for all SKUs.

Note that during the model training and validation phase, we selected the trained policy with ($\lambda_{s1}^*$ ($c_1^{\lambda_{s1}^*}$ =10, $c_2^{\lambda_{s1}^*}$=10, $f_{\text{ref}}^{\lambda_{s1}^*}$ = 0.3) as optimal for Situation 1 (S1), and the trained policy with $\lambda_{s2}^*$ ($c_1^{\lambda_{s2}^*}$ =10, $c_2^{\lambda_{s2}^*}$=10, $f_{\text{ref}}^{\lambda_{s2}^*}$ = 0.5) as the optimal choice for Situation 2 (S2).

Then, using the testing dataset, we performed probabilistic evaluation simulations for the two selected optimal policies (50 Monte-Carlo Runs). We illustrate the outcome imbalance distributions (scaled per SKU by historical maximum inventory, positive values for excess stock, negative values for lost sales) at the 13th timestep, the % excess stock over the 13 timesteps, % lost sales over the 13 timesteps, and the total costs for each situation at the 13th timestep, in comparison to the historical 100% baselines in Table 3. Note that the outcome imbalance distribution in gray color illustrates the historical distribution. The Mean and SD are calculated over 50 Monte-Carlo Runs.

Table 3: Testing Evaluation Performances for the Selected Optimal Policies ($\lambda_{s1}^*$ and $\lambda_{s2}^*$).

| Policy (risk preference) | Outcome Imbalance Distributions (at the 13th timestep) x-axis: outcome imbalance y-axis: frequency | % Excess Stock (over the 13 timesteps) 100% = historical level | % Out Of Stock (over the 13 timesteps) 100% = historical level | % Total Cost Situation 1 (at the 13th timestep) | % Total Cost Situation 2 (at the 13th timestep) |
|---|---|---|---|---|---|
| $c_1^{\lambda_{s1}^*}=10$, $c_2^{\lambda_{s1}^*}=10$, $f_{ref}^{\lambda_{s1}^*}=0.3$ | | Mean = 80%; SD = 5% | Mean = 25%; SD = 7.3% | **Mean = 52%** **SD = 6.1%** | Mean = 34% SD = 7% |
| $c_1^{\lambda_{s2}^*}=10$, $c_2^{\lambda_{s2}^*}=10$, $f_{ref}^{\lambda_{s2}^*}=0.5$ | | Mean = 96%; SD = 5.4% | Mean = 19%; SD = 5.4% | Mean = 57% SD = 5.4% | **Mean = 32%** **SD = 5.4%** |

Furthermore, even though evaluating the testing performance of sub-optimal policies from the validation phase is not required, we offer our analysis for illustrative purposes. We conducted simulations using all other risk-preferential policies on the testing dataset and compared their performances to the optimal policies in Table 4.

Table 4: Testing Evaluation Performance for All Risk-Preferential Policies.

| Policy (risk preference) | | Outcome Imbalance Distributions (at the 13th timestep) x-axis: outcome imbalance y-axis: frequency | % Excess Stock (over the 13 timesteps) 100% = historical level | % Out Of Stock (over the 13 timesteps) 100% = historical level | % Total Cost Situation 1 (at the 13th timestep) | % Total Cost Situation 2 (at the 13th timestep) |
|---|---|---|---|---|---|---|
| $\lambda=1$ | $c_1^\lambda=10$, $c_2^\lambda=10$, $f_{ref}^\lambda=0.0$ | | Mean = 71%; SD = 4.1% | Mean = 74%; SD = 22.6% | Mean = 72% SD = 13.4% | Mean = 74% SD = 19.6% |
| $\lambda=2$ | $c_1^\lambda=10$, $c_2^\lambda=10$, $f_{ref}^\lambda=0.1$ | | Mean = 69%; SD = 3.2% | Mean = 45%; SD = 12.7% | Mean = 57% SD = 8% | Mean = 49% SD = 11.1% |
| $\lambda=3$ | $c_1^\lambda=10$, $c_2^\lambda=10$, $f_{ref}^\lambda=0.2$ | | Mean = 81%; SD = 5.5% | Mean = 25%; SD = 8.7% | Mean = 53% SD = 7.1% | Mean = 34% SD = 8.2% |
| $\lambda=4$ (Selected Optimal for Situation 1, $\lambda_{s1}^*$ in Validation) | $c_1^\lambda=10$, $c_2^\lambda=10$, $f_{ref}^\lambda=0.3$ | | Mean = 80%; SD = 5% | Mean = 25%; SD = 7.3% | **Mean = 52%** **SD = 6.1%** | Mean = 34% SD = 7% |
| $\lambda=5$ | $c_1^\lambda=10$, $c_2^\lambda=10$, $f_{ref}^\lambda=0.4$ | 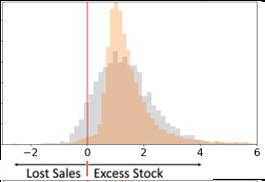 | 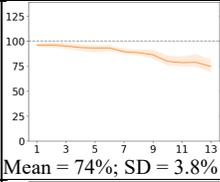 Mean = 74%; SD = 3.8% | 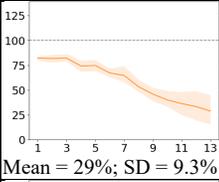 Mean = 29%; SD = 9.3% | Mean = 52% SD = 6.6% | Mean = 36% SD = 8.4% |
| $\lambda=6$ (Selected Optimal for Situation 2, $\lambda_{s2}^*$ in Validation) | $c_1^\lambda=10$, $c_2^\lambda=10$, $f_{ref}^\lambda=0.5$ | 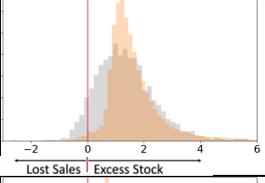 | 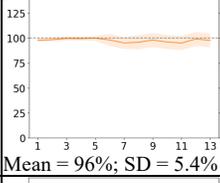 Mean = 96%; SD = 5.4% | 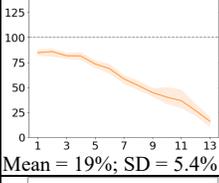 Mean = 19%; SD = 5.4% | Mean = 57% SD = 5.4% | **Mean = 32% SD = 5.4%** |
| $\lambda=7$ | $c_1^\lambda=2$, $c_2^\lambda=10$, $f_{ref}^\lambda=0.0$ | 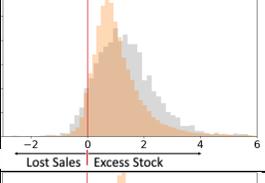 | 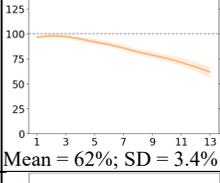 Mean = 62%; SD = 3.4% | 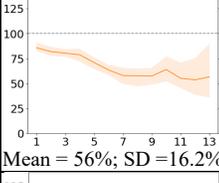 Mean = 56%; SD = 16.2% | Mean = 59% SD = 10.9% | Mean = 57% SD = 14.1% |
| $\lambda=8$ | $c_1^\lambda=2$, $c_2^\lambda=10$, $f_{ref}^\lambda=0.1$ | 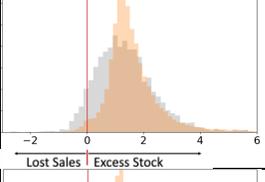 | 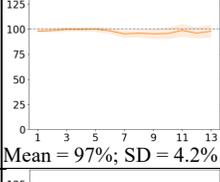 Mean = 97%; SD = 4.2% | 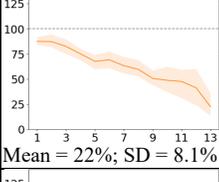 Mean = 22%; SD = 8.1% | Mean = 60% SD = 6.1% | Mean = 35% SD = 7.5% |
| $\lambda=9$ | $c_1^\lambda=2$, $c_2^\lambda=10$, $f_{ref}^\lambda=0.2$ | 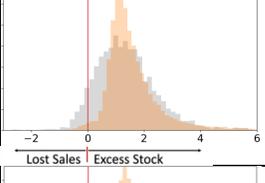 | 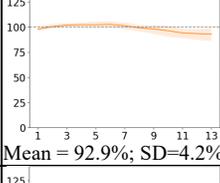 Mean = 92.9%; SD=4.2% | 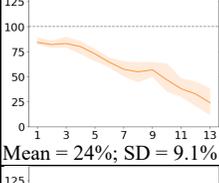 Mean = 24%; SD = 9.1% | Mean = 58% SD = 6.6% | Mean = 35.5% SD = 8.3% |
| $\lambda=10$ (Selected Second Best for Situation 2 in Validation) | $c_1^\lambda=2$, $c_2^\lambda=10$, $f_{ref}^\lambda=0.3$ | 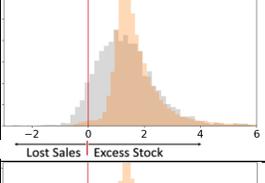 | 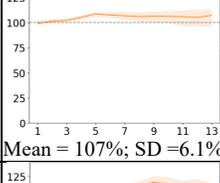 Mean = 107%; SD =6.1% | 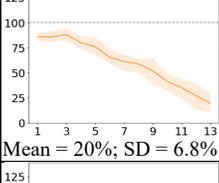 Mean = 20%; SD = 6.8% | Mean = 62% SD = 6.7% | Mean = 34% SD = 6.7% |
| $\lambda=11$ | $c_1^\lambda=2$, $c_2^\lambda=10$, $f_{ref}^\lambda=0.4$ | 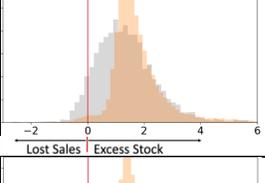 | 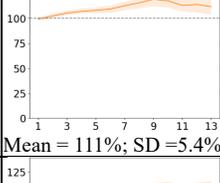 Mean = 111%; SD =5.4% | 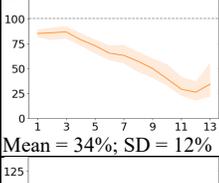 Mean = 34%; SD = 12% | Mean = 73% SD = 8.7% | Mean = 47% SD = 10.9% |
| $\lambda=12$ | $c_1^\lambda=2$, $c_2^\lambda=10$, $f_{ref}^\lambda=0.5$ | 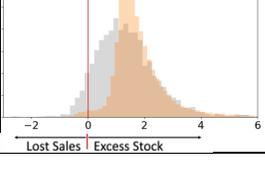 | 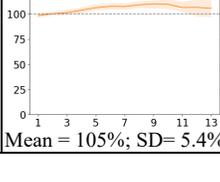 Mean = 105%; SD= 5.4% | 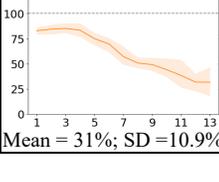 Mean = 31%; SD =10.9% | Mean = 68% SD = 8.1% | Mean = 44% SD = 9.9% |

# E. GPP Training Algorithm

---

**Algorithm 1: GPP Training**

---

**Input**: $\mathcal{G} = (\mathcal{V}, \mathcal{E}), \mathfrak{D} = \{(x^t, a^t, x^{t+1}, a^{t+1}) | t \in \mathbb{N}\}$,
       initialized networks $Q(x, a; \theta_Q)$ and $\mu(x; \theta_\mu)$
       initialized target networks $\theta_Q^{\text{target}} \leftarrow \theta_Q$, $\theta_\mu^{\text{target}} \leftarrow \theta_\mu$

**Parameter**: $c_1^\lambda, c_2^\lambda, f_{\text{ref}}^\lambda, \gamma, \eta, \tau$

**Output**: optimized $\theta_Q$ and $\theta_\mu$

1: **for** t = 1, T **do**
2:     Fetch a minibatch of transitions $(x, a, x', a')$ from $\mathfrak{D}$
3:     Calculate $q = Q(x, a)$ in (6).
4:     **if** t < N    // train Q alone with historical transition samples
5:         $y = r(x, a) + \gamma Q(x', a'; \theta_Q^{\text{target}})$
6:         Update $\theta_Q$ using $\nabla_{\theta_Q} L(\theta_Q)$ in (8).
7:         $\theta_Q^{\text{target}} \leftarrow \tau \theta_Q + (1 - \tau) \theta_Q^{\text{target}}$
8:     **else**    // train both $Q$ and $\mu$
9:         $y = r(x, a) + \gamma Q(x', \mu(x'; \theta_\mu^{\text{target}}); \theta_Q^{\text{target}})$
10:       Update $\theta_Q$ using $\nabla_{\theta_Q} L(\theta_Q)$ in (8).
11:       Update $\theta_\mu$ using $\nabla_{\theta_\mu} J(\theta_\mu)$ in (9).
12:       $\theta_\mu^{\text{target}} \leftarrow \tau \theta_\mu + (1 - \tau) \theta_\mu^{\text{target}}$
13:       $\theta_Q^{\text{target}} \leftarrow \tau \theta_Q + (1 - \tau) \theta_Q^{\text{target}}$
14:     **end if**
15: **end for**
16: **return** $\theta_Q$ and $\theta_\mu$

## F. GPP Planning Algorithm

**Algorithm 2: GPP Planning**

**Input**: $\mathcal{G} = (\mathcal{V}, \mathcal{E})$, trained policy $\mu^\lambda(\mathbf{x}; \theta_\mu)$ for risk preference $\lambda$,
time of planning $t$, initial inventory $I_v^t$,
probabilistic demand prediction model $p(D_v^{j|t})$,
probabilistic lead time prediction model $p(\tau_{wv}^l)$,
probabilistic production prediction model $p(U_v^t)$ or planned production

**Parameter**: time horizon $J$, number of predicted imbalance features $K$,
number of probabilistic Monte-Carlo $Z$,
economic objective cost parameters $c_{es}$ and $c_{oos}$

**Output**: supply action plan over $J$ time intervals

1: initialize $\hat{I}_v^t = I_v^t$
2: **for** $z = 1, Z$ **do**
3:     sampling predicted demand $\{\widehat{D}_v^{j|t} | j \in \{t, \ldots, t + J + K - 3\}\} \sim p(D_v^{j|t})$
4:     sampling predicted lead time $\{\hat{\tau}_{wv}^{t'} | t' \in \{0, \ldots, t + J - 1\}\} \sim p(\tau_{wv}^{t'})$ or historical
5:     sampling production $\{\widehat{U}_v^{j|t} | j \in \{t, \ldots, t + J + K - 3\}\} \sim p(U_v^{j|t})$ or plan
6:     **for** $j = t, t + J - 1$ **do**
7:         $f_v^{j|j} = \hat{I}_v^j$
8:         **for** $k = 1, K - 1$ **do**
9:             **for** $v \in$ distribution nodes **do**
10:                 $\hat{S}_v^{j+k-1|j} = \sum_{w \in \mathcal{N}_{src}(v)} \sum_{m=1}^M \sum_{t'=0}^{j-1} a_{wv}^{t'}[m] \, \delta\{\hat{\tau}_{wv}^{t'}[m] = j + k - 1 - t'\}$
11:                 $f_v^{j+k|j} = f_v^{j+k-1|j} + \hat{S}_v^{j+k-1|j} - \widehat{D}_v^{j+k-1|t}$
12:             **end for**
13:             **for** $v \in$ production nodes **do**
14:                 $f_v^{j+k|j} = f_v^{j+k-1|j} + \widehat{U}_v^{j+k-1|j}$
15:             **end for**
16:         **end for**
17:         $x_v^j = (f_v^{j|j}, f_v^{j+1|j}, \ldots, f_v^{j+K|j})$
18:         $\mathbf{x}^j = \{x_v^j \in \mathbb{R}^K : v \in \mathcal{V}\}$
19:         $\mathbf{a}^j = \mu^\lambda(\mathbf{x}^j)$
20:         $A_v^j = \sum_{w \in \mathcal{N}_{dst}(v)} \sum_{m=1}^M a_{vw}^j[m]$
21:         **for** $v \in$ distribution nodes **do**
22:             $\hat{S}_v^{j|j+1} = \sum_{w \in \mathcal{N}_{src}(v)} \sum_{m=1}^M \sum_{t'=0}^{j} a_{wv}^{t'}[m] \, \delta\{\hat{\tau}_{wv}^{t'}[m] = j - t'\}$
23:             $\hat{I}_v^{j+1} = \hat{I}_v^j + \hat{S}_v^{j|j+1} - \widehat{D}_v^{j|t} - A_v^j$
24:         **end for**
25:         **for** $v \in$ production nodes **do**
26:             $\hat{I}_v^{j+1} = \hat{I}_v^j + \widehat{U}_v^{j|t} - A_v^j$
27:         **end for**
28:         **for** $v \in$ distribution nodes **do**
29:             $OOS_v^j = \min\{\hat{I}_v^j + \hat{S}_v^{j|j+1} - D_v^{j|t}, 0\}$
30:             $ES_v^j = \max\{\hat{I}_v^{j+1}, 0\}$
31:             $Cost_v^j = c_{es} |ES_v^t| + c_{oos} |OOS_v^t|$
32:         **end for**
33:         $Cost^j = \sum_v Cost_v^j$
34:     **end for**
35:     $Cost_z = (1/J) \sum_j Cost^j$
36:     $\mathcal{A}_z = \{\mathbf{a}^j | j \in \{t, \ldots, t + J - 1\}\}$
37: **end for**

38: $AvgCost_\lambda(t) = (1/Z) \sum_z Cost_z$
39: $\{\bar{a}^{\lambda,j} | j \in \{t, \ldots, t+J-1\}\} = (1/Z) \sum_z \mathcal{A}_z$
40: **return** $\{\bar{a}^{\lambda,j} | j \in \{t, \ldots, t+J-1\}\}, AvgCost_\lambda(t)$

---

We employ probabilistic Monte-Carlo (MC) for generating the action plan via one or more trained policies, each with varying risk preferences. Even with the same initial inventory, incorporating probabilistic future demands, lead times, and production conditions results in a diverse set of MC sampled states representing the current probabilistic network state. For each policy, we compute actions and costs across the time horizon from each MC state, using the economic costs of the current objective. The optimal policy is chosen based on the lowest averaged cost over all MC states. The final planned actions are determined by averaging sampled actions from the selected policy.

In the algorithm above, at every timestep $t$, we choose $\lambda^*(t) = \arg\min_\lambda AvgCost_\lambda(t)$ for the specific objective described by cost parameters $c_{es}$ and $c_{oos}$ at the time of planning. Then the optimal plan $\{\bar{a}^{\lambda^*,j} | j \in \{t, \ldots, t+J-1\}\}$ generated from $\mu^{\lambda^*}(x; \theta_\mu)$ is recommended for the upcoming time horizon. As a simple alternative, we could employ the optimal policy selected during the model validation phase for the cost objective.

## G. Evaluation of a Rule-Based Policy

As another point of comparison, we carried out simulations using a rule-based policy where each node calculates SKU/node-specific safety stock based on predicted total demand using the safety days of supply (DOS) information that the company provided. The rule-based policy operates through the following steps:

1. Each node calculates SKU/node-specific safety stock using predicted total demand for safety days of supply (DOS).
2. Each node asks a parent node, selected randomly according to historical supply proportions among its parents, to provide supply equal to the imbalance (= max (safety stock − inventory, 0)) at each timestep.
3. The parent node's ability to provide supply to its child nodes is limited by the remaining stock after fulfilling its own demand.

In the evaluation of testing performance, this fixed rule-based policy (which does not adapt to cost objectives) is considerably deficient, leading to a significant 166% increase (averaged over 50 Monte-Carlo runs, SD = 3%) in lost sales, despite achieving a 36% reduction[3] (averaged over 50 Monte-Carlo runs, SD = 1%) in excess stock (compared to historical 100% level). This highlights the complexity of the underlying supply chain planning challenge.

## H. Computing Setup and Runtime

Our code was executed in Python 3.8.3 using the following environment:
- pandas=1.1
- numpy=1.22
- scikit-learn=1.1
- pytorch=1.12
- torch_geometric=2.3

Training and evaluation were executed in a composable Kubeflow pipeline on a scalable Kubernetes cluster which permitted us to run multiple experiments in parallel and has the capability to execute different steps of the workflow independently. This allowed us to use different computational resources depending on what was needed on each of the steps. On data processing steps that didn't require a high number of resources, we used instances with 4 vCPUs and 32 GiB of memory. When we were required to run memory-intense data processing steps (e.g., preparing network snapshots and running simulations), we used bigger instances that have 32 vCPUs and 64 GiB of memory. For the training of the models, we used GPU instances with a GPU memory of 24 GiB and 16 vCPUs.

We ran multiple experiments using a broad set of hyperparameters. Each run took an average of 8 hours to complete; where ~1 hour on data processing, ~5 hours were on training and evaluation, and ~2 hours on simulations.

---

[3] There was a typographical error in our main paper where a 60% reduction was maintained. The correct value is a 36% reduction. We will rectify this in our final publication.

# I. Visualizing Generative Supply Chain Networks in GPP Simulations

GPP can *generate* probabilistic samples of the dynamic evolution of supply chain networks with supply actions that optimize costs for specific objectives, considering probabilistic current states and future conditions inferred from current inventory, predicted demand, lead time and production. We demonstrate GPP-generated optimal actions over time (13 weeks) through map-based supply chain network visualization, compared with the historical actions.

Figure 8 outlines the viewing instructions for the *Generative Supply Chain Network Visualization*.

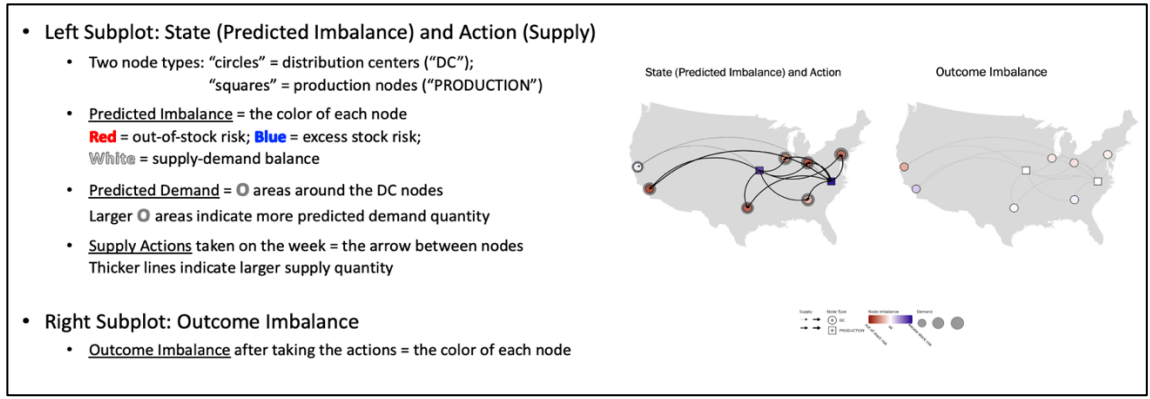

Figure 8: Viewing Instructions for Generative Supply Chain Network Visualization

Figure 9 demonstrates an example week of network transition snapshots comparing GPP with historical actions in the map-based visualization. Kindly review **the enclosed PDF titled "Visualizing Generative Supply Chain Networks in GPP Simulations"** for complete simulation visual snapshots over a span of 13 weeks, starting from a network state of a specific SKU/week in a testing evaluation period.

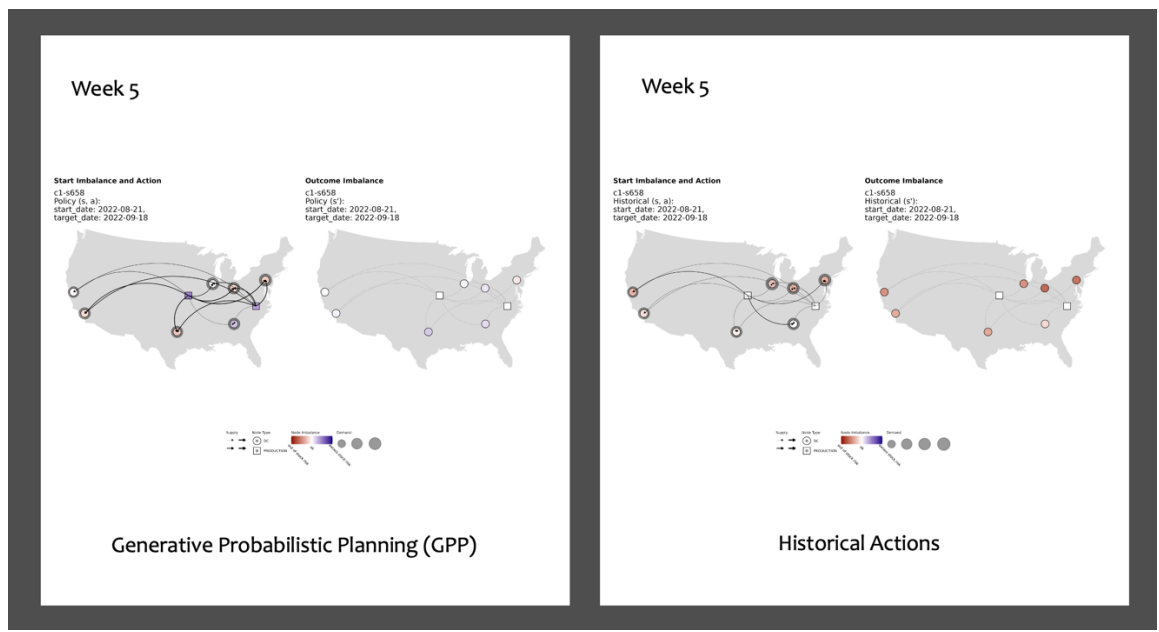

Figure 9: Comparing GPP with Historical Actions on Week 5 (Example).

The key takeaways from the Generative Supply Chain Network Visualization are summarized in Figure 10.

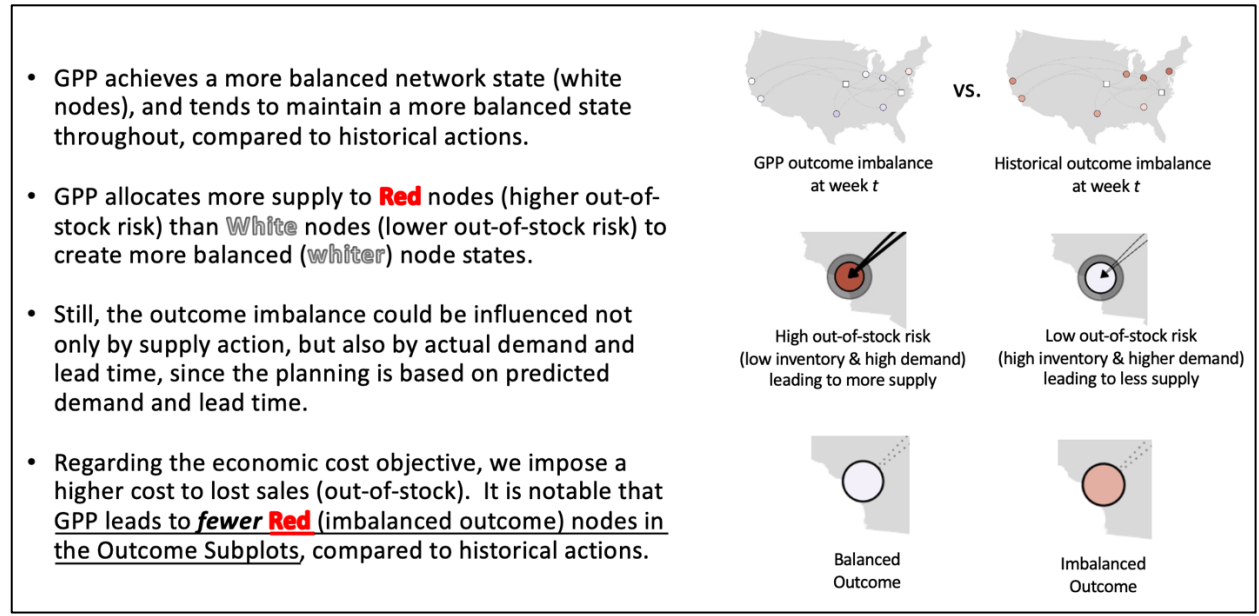

- GPP achieves a more balanced network state (white nodes), and tends to maintain a more balanced state throughout, compared to historical actions.

- GPP allocates more supply to **Red** nodes (higher out-of-stock risk) than White nodes (lower out-of-stock risk) to create more balanced (whiter) node states.

- Still, the outcome imbalance could be influenced not only by supply action, but also by actual demand and lead time, since the planning is based on predicted demand and lead time.

- Regarding the economic cost objective, we impose a higher cost to lost sales (out-of-stock). It is notable that GPP leads to *fewer* **Red** (imbalanced outcome) nodes in the Outcome Subplots, compared to historical actions.

Figure 10: Key Insights from Generative Supply Chain Network Visualization

# Technical Appendix

Visualizing Generative Supply Chain Networks
in GPP Simulations

# Viewing Instructions for Generative Supply Chain Network Visualization

- Left Subplot: State (Predicted Imbalance) and Action (Supply)
    - Two node types: "circles" = distribution centers ("DC");
      "squares" = production nodes ("PRODUCTION")
    - Predicted Imbalance = the color of each node
      Red = out-of-stock risk; Blue = excess stock risk;
      White = supply-demand balance
    - Predicted Demand = O areas around the DC nodes
      Larger O areas indicate more predicted demand quantity
    - Supply Actions taken on the week = the arrow between nodes
      Thicker lines indicate larger supply quantity

- Right Subplot: Outcome Imbalance
    - Outcome Imbalance after taking the actions = the color of each node

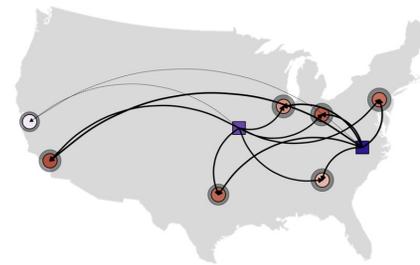
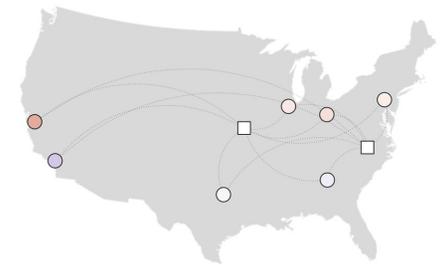
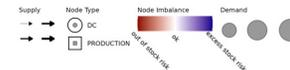

# Key Insights from Generative Supply Chain Network Visualization

- GPP achieves a more balanced network state (white nodes), and tends to maintain a more balanced state throughout, compared to historical actions.

- GPP allocates more supply to **Red** nodes (higher out-of-stock risk) than White nodes (lower out-of-stock risk) to create more balanced (whiter) node states.

- Still, the outcome imbalance could be influenced not only by supply action, but also by actual demand and lead time, since the planning is based on predicted demand and lead time.

- Regarding the economic cost objective, we impose a higher cost to lost sales (out-of-stock). It is notable that GPP leads to *fewer* **Red** (imbalanced outcome) nodes in the Outcome Subplots, compared to historical actions.

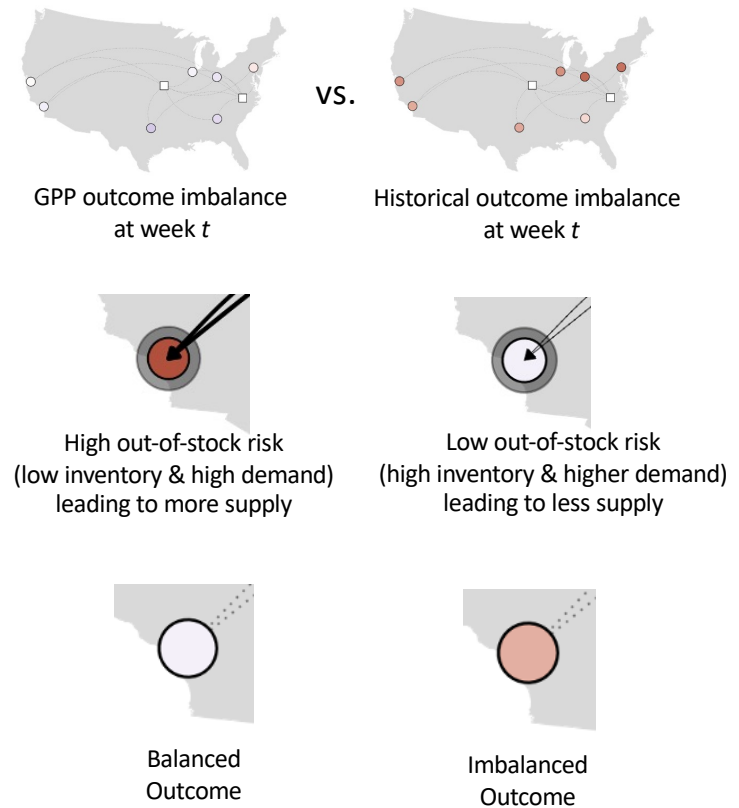

GPP outcome imbalance at week *t* vs. Historical outcome imbalance at week *t*

High out-of-stock risk (low inventory & high demand) leading to more supply

Low out-of-stock risk (high inventory & higher demand) leading to less supply

Balanced Outcome

Imbalanced Outcome

Presented below are visual representations depicting the weekly network transition snapshots (state-action, outcome) spanning 13 weeks, originating from a network state of a specific SKU/week in a testing evaluation period.

The left panel showcases GPP, while the right panel displays historical actions for comparison.

Week 1

State (Predicted Imbalance) and Action 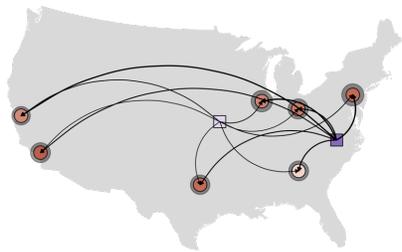 Outcome Imbalance 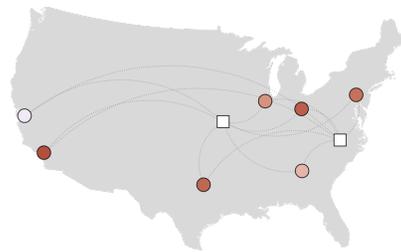

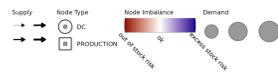

Generative Probabilistic Planning (GPP)

Week 1

State (Predicted Imbalance) and Action 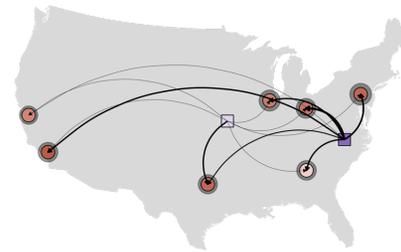 Outcome Imbalance 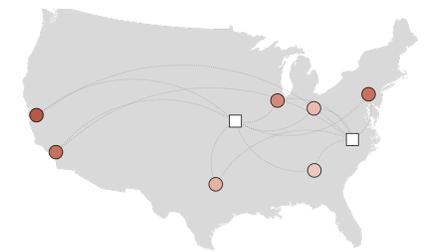

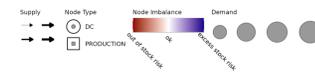

Historical Actions

Week 2

State (Predicted Imbalance) and Action
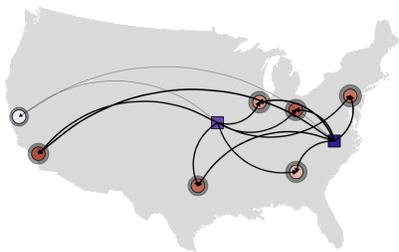

Outcome Imbalance
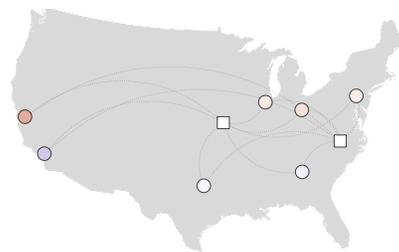

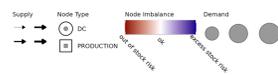

Generative Probabilistic Planning (GPP)

Week 2

State (Predicted Imbalance) and Action
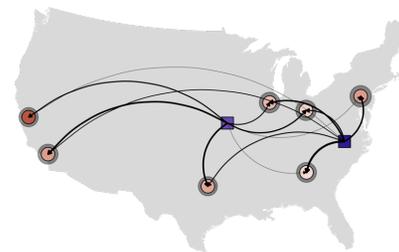

Outcome Imbalance
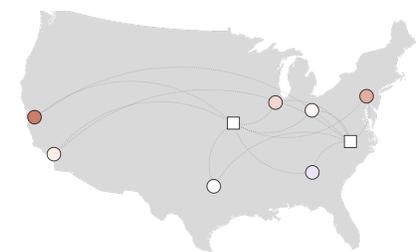

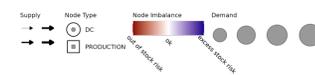

Historical Actions

Week 3

State (Predicted Imbalance) and Action     Outcome Imbalance

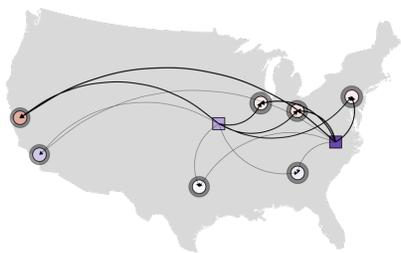 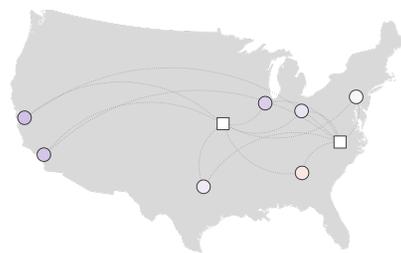

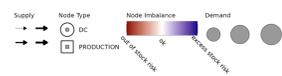

Generative Probabilistic Planning (GPP)

Week 3

State (Predicted Imbalance) and Action     Outcome Imbalance

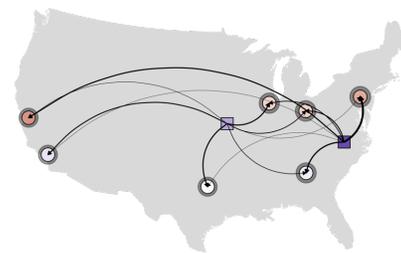 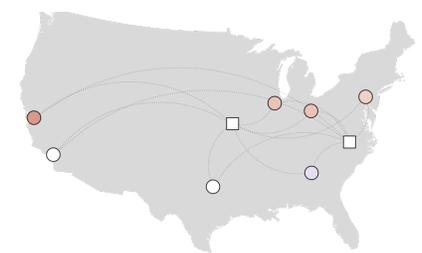

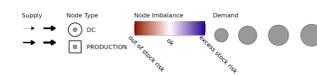

Historical Actions

Week 4

State (Predicted Imbalance) and Action    Outcome Imbalance

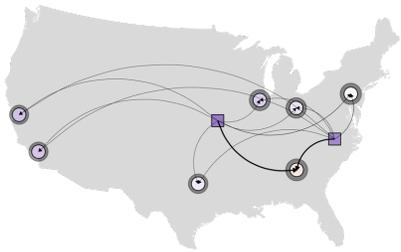
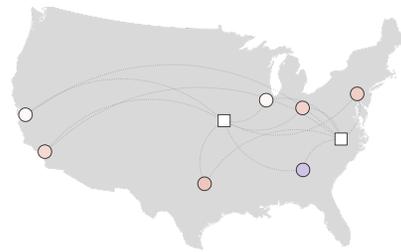

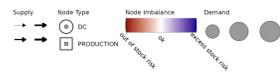

Generative Probabilistic Planning (GPP)

Week 4

State (Predicted Imbalance) and Action    Outcome Imbalance

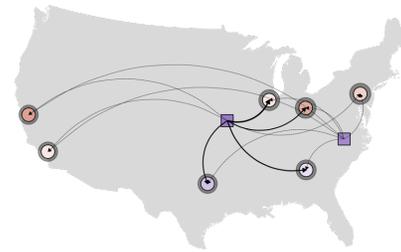
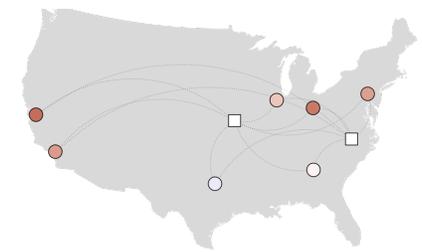

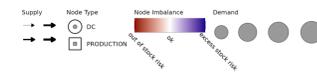

Historical Actions

Week 5

State (Predicted Imbalance) and Action 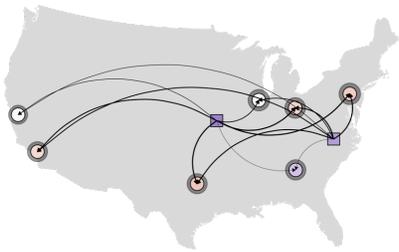   Outcome Imbalance 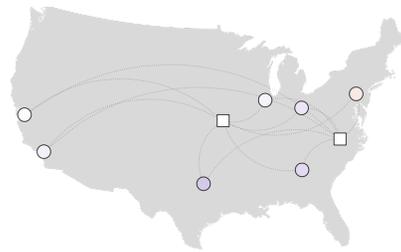

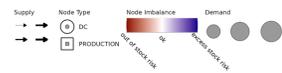

Generative Probabilistic Planning (GPP)

Week 5

State (Predicted Imbalance) and Action 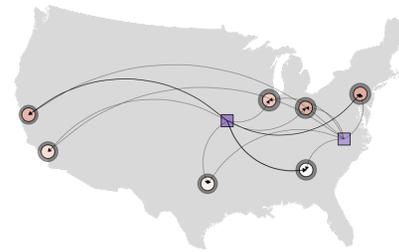   Outcome Imbalance 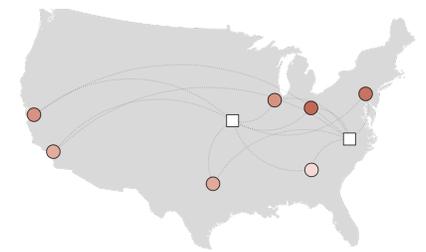

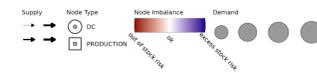

Historical Actions

Week 6

State (Predicted Imbalance) and Action 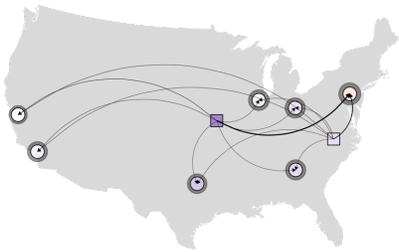  Outcome Imbalance 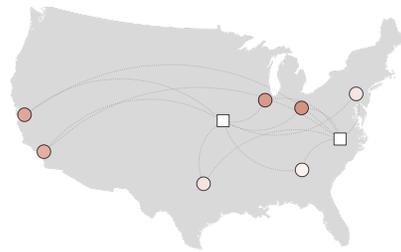

Generative Probabilistic Planning (GPP)

Week 6

State (Predicted Imbalance) and Action 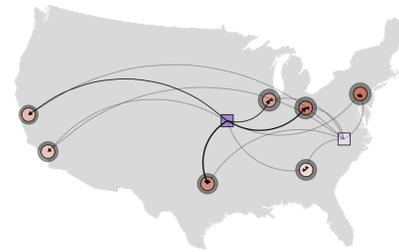  Outcome Imbalance 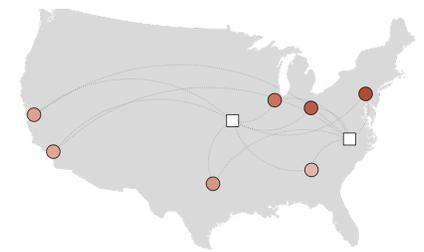

Historical Actions

Week 7

State (Predicted Imbalance) and Action 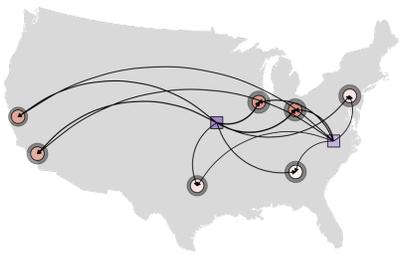  Outcome Imbalance 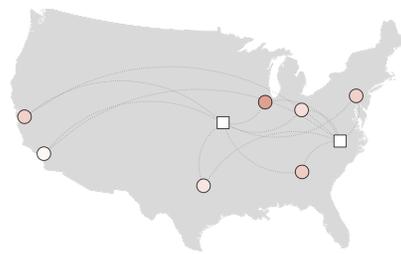

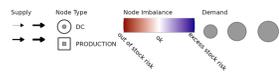

Generative Probabilistic Planning (GPP)

Week 7

State (Predicted Imbalance) and Action 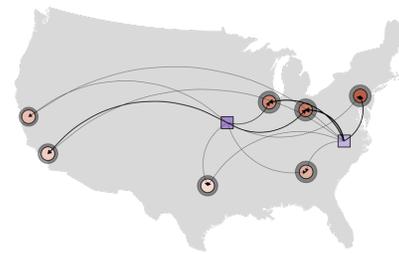  Outcome Imbalance 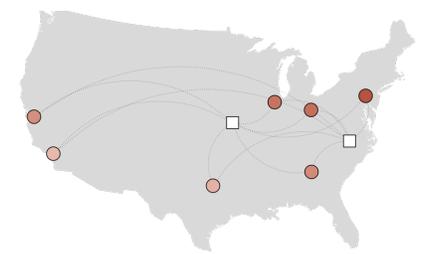

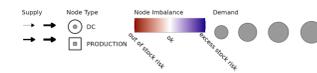

Historical Actions

Week 8

State (Predicted Imbalance) and Action    Outcome Imbalance

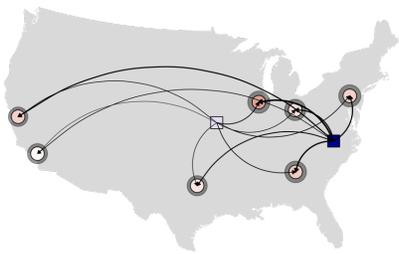 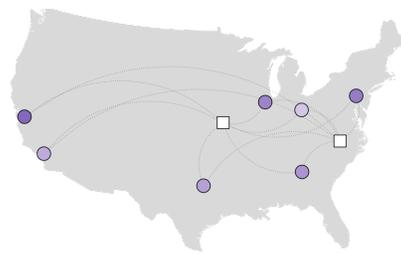

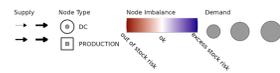

Generative Probabilistic Planning (GPP)

Week 8

State (Predicted Imbalance) and Action    Outcome Imbalance

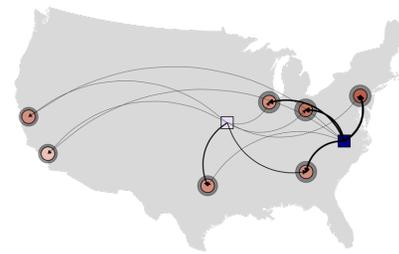 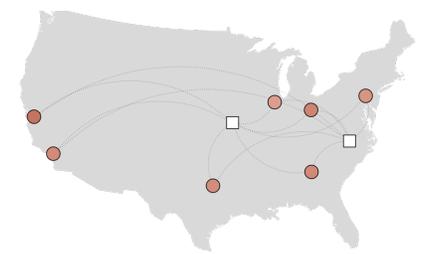

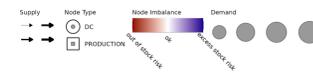

Historical Actions

Week 9

State (Predicted Imbalance) and Action 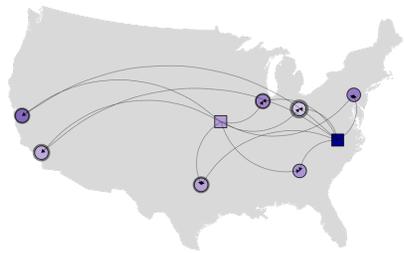 Outcome Imbalance 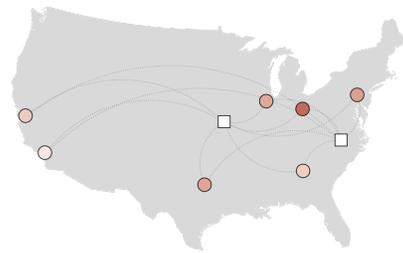

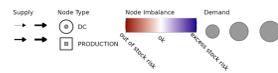

Generative Probabilistic Planning (GPP)

Week 9

State (Predicted Imbalance) and Action 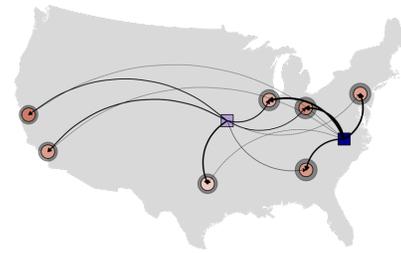 Outcome Imbalance 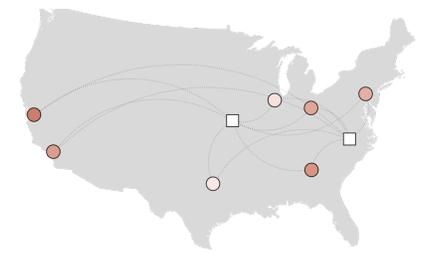

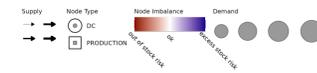

Historical Actions

Week 10

State (Predicted Imbalance) and Action    Outcome Imbalance

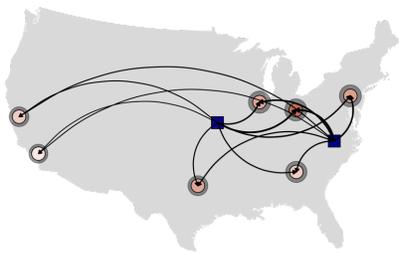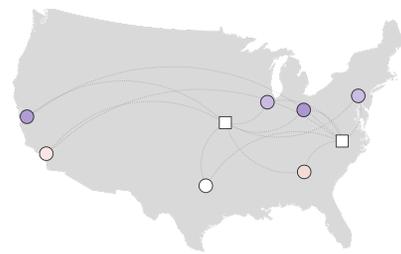

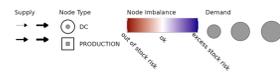

Generative Probabilistic Planning (GPP)

Week 10

State (Predicted Imbalance) and Action    Outcome Imbalance

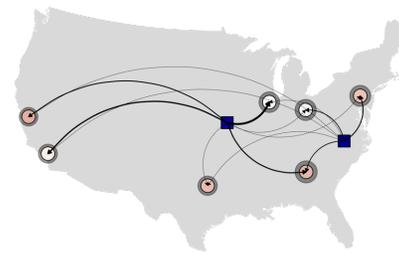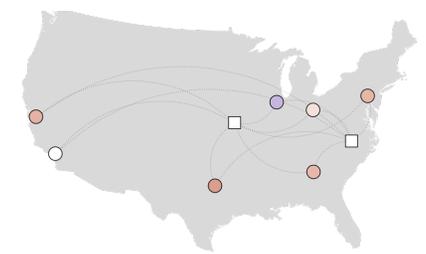

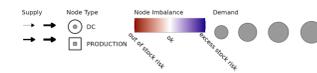

Historical Actions

Week 11

State (Predicted Imbalance) and Action  Outcome Imbalance

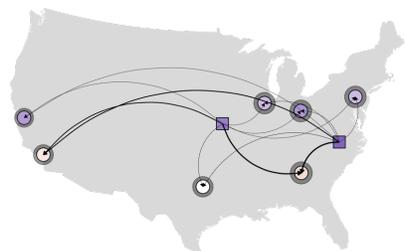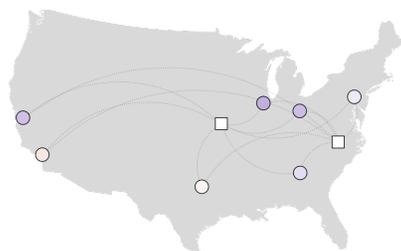

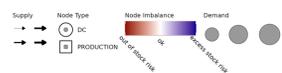

Generative Probabilistic Planning (GPP)

Week 11

State (Predicted Imbalance) and Action  Outcome Imbalance

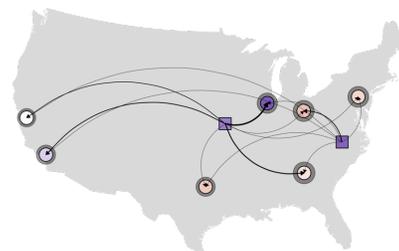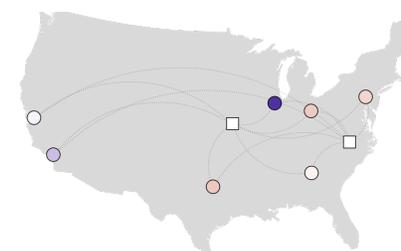

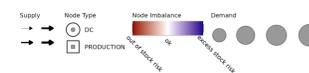

Historical Actions

Week 12

State (Predicted Imbalance) and Action 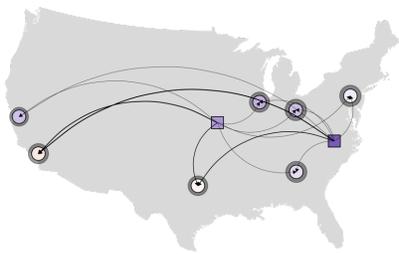
Outcome Imbalance 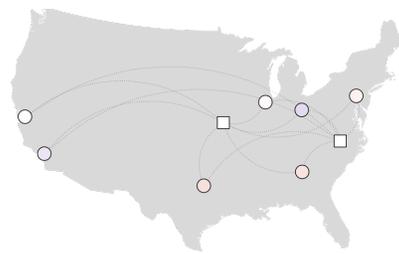

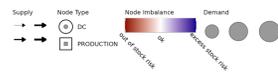

Generative Probabilistic Planning (GPP)

Week 12

State (Predicted Imbalance) and Action 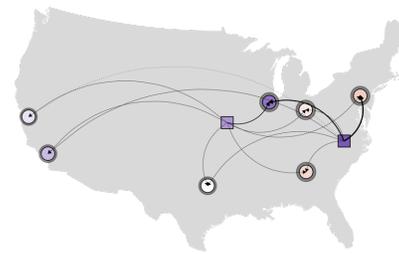
Outcome Imbalance 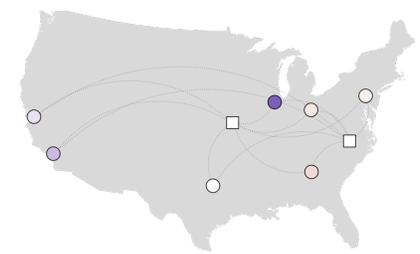

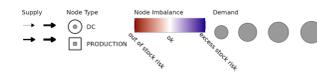

Historical Actions

Week 13

State (Predicted Imbalance) and Action    Outcome Imbalance

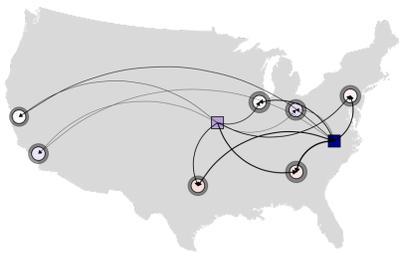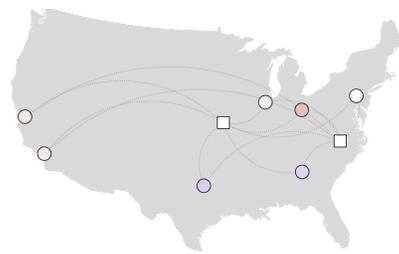

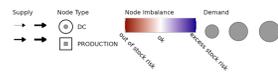

Generative Probabilistic Planning (GPP)

Week 13

State (Predicted Imbalance) and Action    Outcome Imbalance

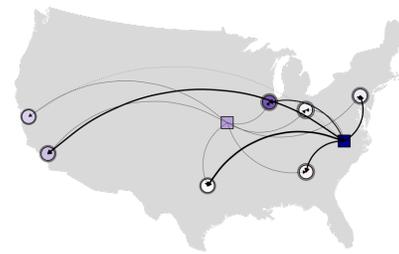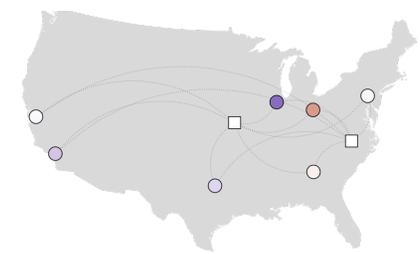

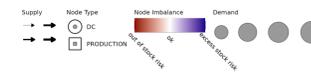

Historical Actions